\documentclass{article}

\usepackage[preprint]{neurips_2026}

\usepackage[utf8]{inputenc}
\usepackage[T1]{fontenc}
\usepackage{hyperref}
\usepackage{url}
\usepackage{booktabs}
\usepackage{graphicx}
\usepackage{placeins}
\usepackage{longtable}
\usepackage{amsfonts}
\usepackage{amsmath}
\usepackage{nicefrac}
\usepackage{microtype}
\usepackage{xcolor}
\usepackage[capitalise]{cleveref}

\title{Unifying matrix completion across domains with a synthetically pretrained foundation model}

\author{%
  Dwaipayan Saha$^\star$ \\
  Columbia University \\
  \And
  Jacob Feitelberg$^\star$ \\
  Columbia University \\
  \And
  Kyuseong Choi$^\star$ \\
  Cornell Tech \\
  \And
  Anish Agarwal \\
  Columbia University, Traversal
    \And
  Raaz Dwivedi \\
  Cornell Tech, Traversal
}

\usepackage{float}
\begin{document}

\maketitle

\begin{abstract}
  Matrix completion underlies problems from tabular imputation to causal inference. While matrix completion provides a common setup for these domains, the best methods in each domain are still specialized and do not transfer well across domains. We introduce MatrixFormer, a pre-trained matrix-native transformer that predicts a full distribution for every missing entry in a single forward pass. MatrixFormer is trained entirely on synthetic low-rank and latent-factor matrices under diverse missingness patterns. Applied zero-shot with fixed network weights, MatrixFormer achieves competitive performance on causal inference panel-data tasks, tabular imputation, and recommendation systems matrix completion. These results position MatrixFormer as a general-purpose foundation model for matrix completion.
\end{abstract}

\section{Introduction}
Matrix completion is the task of imputing missing entries in a partially observed numerical matrix. This common task appears in areas across both science and business: predicting user ratings \citep{salah2020cornac}, imputing tabular missing data \citep{jarrett2022hyperimpute}, and estimating counterfactual outcomes in panel data \citep{athey2021matrix}. However, there is a tension currently in the center of this field: while the problem formulation is shared across many domains, the state-of-the-art methods used in each domain are not shared. We propose a single learned foundation model to close this gap. We open-source our training and evaluation code\footnote{\url{https://github.com/jacobf18/MatrixFormer}}, as well the model\footnote{\url{https://huggingface.co/Tabimpute/MatrixFormer}}.

Classical general matrix completion methods such as SoftImpute \citep{mazumder2010spectral} rely on specific assumptions to work such as the matrix having a low rank and entries being missing at random. Matrices from different domains have very different data-generating processes and missingness patterns: a ratings matrix is almost completely missing and is bounded from 1-5 or 1-10, panel data has contiguous missing blocks based on entry treatment patterns, and tabular data often has different scales in each column with biased random missingness. Each domain violates the standard assumptions for the general methods in different ways, which results in a loss in accuracy.

As a result of each domain drifting from standard matrix completion assumptions, researchers have proposed specialized methods that take advantage of the different data and missingness patterns in each domain. BellKor’s Pragmatic Chaos \citep{koren2009bellkor}, which won the \$1 million Netflix Prize \citep{bennett2007netflix} for recommendation systems, built highly specialized tools to take advantage of user and item biases not present in standard assumptions. In causal inference with panel data, synthetic difference-in-differences \citep{arkhangelsky2021synthetic} have been shown to outperform standard matrix completion methods to predict causal effects, but are based on a block treatment pattern and reweighted parallel trends, neither of which immediately relate to ratings or tabular data.

While specialization improved performance on each domain, it also splintered the field where advancements in one domain no longer translated to the rest. Practitioners must know which regime their matrix is in and apply the correct specialized method. On top of this, for any new applications of matrix completion, researchers must start from scratch to build up good specialized methods.

To remedy these problems, we introduce \textbf{MatrixFormer}, a foundation model for matrix completion. MatrixFormer is transformer ($\sim$78M parameters) pretrained only on over a million synthetic partially-observed matrices drawn from a diverse set of data-generating and missingness priors. A single forward pass on the matrix produces accurate imputations with no fine-tuning necessary. A general method works here because it does not assume a single structure for all matrices: instead, it learns many structures from the prior mixture and infers which to use at inference time.

\begin{figure}
    \centering
    \includegraphics[width=0.95\linewidth]{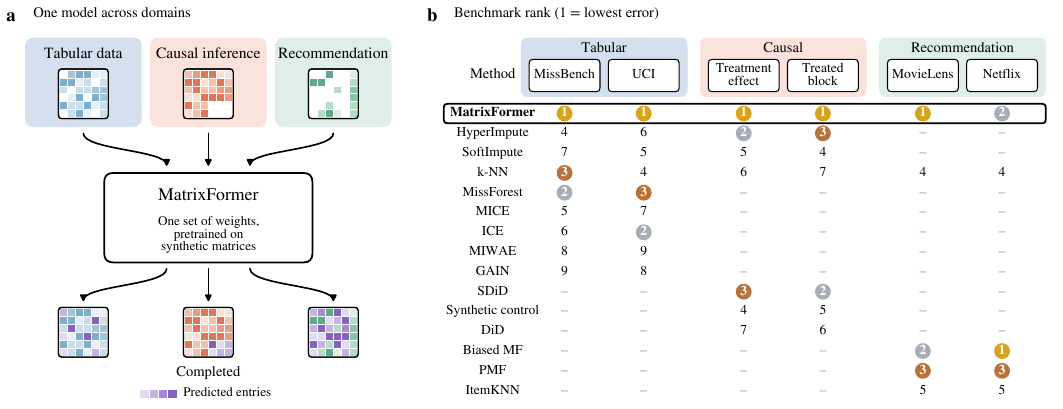}
    \caption{\textbf{Overview of MatrixFormer:} (a) MatrixFormer is pretrained transformer to make accurate predictions across a diverse set of domains. (b) Our method achieves top performance across 3 diverse areas against domain-specific methods.}
    \label{fig:overview}
\end{figure}

Per-domain, MatrixFormer achieves competitive results, as shown in \cref{fig:overview}: (1) for tabular missing-data imputation, MatrixFormer beats MissForest and HyperImpute, (2) for causal inference, MatrixFormer beats synthetic difference-in-differences, and (3) for ratings predictions, MatrixFormer is at-par with a recommendation system-specific matrix factorization method. Finally, extending MatrixFormer's performance to a new domain with data-generating and missingness priors very dissimilar to the ones we used would just require adding in a new prior to our mixture and training the model further, which is much simpler than designing a new algorithm from the ground up.

\subsection{Related Work}
\label{sec:related-work}

\enlargethispage{\baselineskip}
\paragraph{Matrix completion across domains.} Here, we discuss several domains where matrix completion methods are popular. Tabular imputation recovers missing measurements in a sample-feature table. Standard methods such as MICE \citep{vanbuuren2011mice} and HyperImpute \citep{jarrett2022hyperimpute} do so by iteratively modeling each variable conditional on the others. In recommendation systems, the matrix records user-item ratings, and collaborative filtering predicts unobserved ratings from similar users or shared latent factors~\citep{resnick1994grouplens,salakhutdinov2008pmf}. Causal panel studies involve a different kind of missingness. Here, counterfactual outcomes are modeled as missing values in a unit-time matrix. Synthetic controls construct a weighted combination of comparison units~\citep{abadie2010synthetic}, while difference-in-differences and synthetic difference-in-differences use comparisons across units and time~\citep{arkhangelsky2021synthetic}. \citet{athey2021matrix} make the connection to low-rank completion explicit, accommodating the structured missingness of causal panels. The shared setup of these problems motivates a common completion model despite their vast differences in data and missingness patterns.

\paragraph{Prior-data fitted networks.}

Prior-data fitted networks (PFNs), first introduced by \citet{muller2022pfn}, are neural networks trained to approximate posterior predictive distributions using synthetic data generated from prespecified priors. TabPFN \citep{hollmann2025tabpfn} and TabICL \citep{qu2025tabicl} are popular PFNs for supervised learning. TabImpute \citep{feitelberg2025tabimpute} is a PFN built for small-scale tabular imputation. Our work, MatrixFormer, extends this approach to priors over spectral structure, structured missingness, and sparse ratings. We apply MatrixFormer to a wide range domains including tabular imputation, recommendation system, and panel causal inference. At inference time, its weights are fixed, and the observed entries supply context for completion.

\section{Architecture and synthetic data}
\label{sec:architecture-synthetic-data}
MatrixFormer is a neural network with a novel transformer architecture so that it can handle matrices of arbitrary size in both dimensions. We pair this matrix-native architecture with synthetic training tasks with varied structure and missingness patterns.

\subsection{Matrix-native architecture}
\label{sec:architecture}

\begin{figure}[t]
    \centering
    \includegraphics[width=0.95\linewidth]{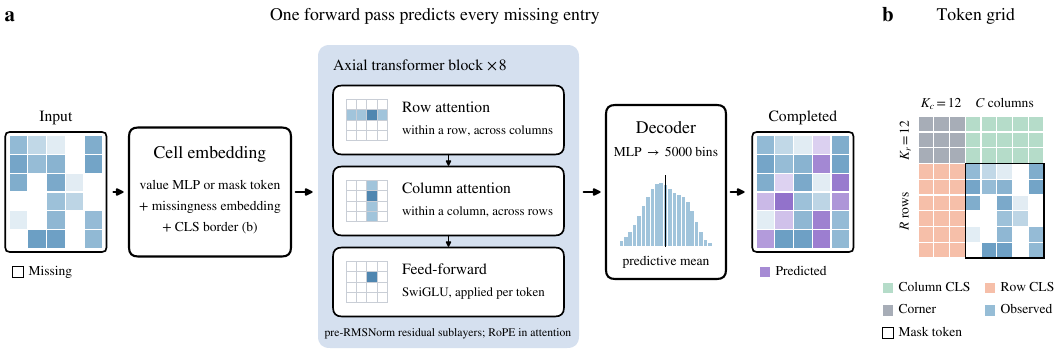}
    \caption{\textbf{MatrixFormer forward pass.} (a) Each cell of the partially observed matrix becomes a token, with a learned mask token at missing entries. Each of the eight transformer blocks applies row attention within each row (feature attention), column attention within each column (datapoint attention), and a feed-forward layer. A decoder maps every token to a predictive distribution over 5000 bins, whose mean fills the missing entry. (b) The token grid: $K_c$ columns of row-CLS tokens and $K_r$ rows of column-CLS tokens border the $R\times C$ data tokens and are removed before decoding.
    }
    \label{fig:matformer-forward}
\end{figure}

We embed each observed matrix cell using a learned embedding layer, map missing entries to learned token, and pad learned tokens (CLS tokens) on both the rows and columns, as shown in \Cref{fig:matformer-forward}. The CLS tokens allow row-wise and column-wise information to be pooled effectively across layers through self-attention \citep{vaswani2017attention}. These tokens give the model a factor-inspired inductive bias without requiring the completed matrix to have a fixed rank. Row and column factors used in collaborative filtering and panel-data models also inform our design~\citep{salakhutdinov2008pmf,athey2021matrix}. In each layer, we cell attends to other cells within its row, followed by a 2nd attention across the column axis, and finally a feed-forward layer. After 8 layers, each token corresponding to a missing entry is decoded into a distributional output similar to TabPFN. We then predict the mean of each distribution as the final prediction for each missing cell. This gives us an architecture tailored to the row-column structure of the data.

\subsection{Synthetic priors}
\label{sec:synthetic-priors}

MatrixFormer is trained entirely on synthetic matrices, so the prior is designed to cover the structure often seen in real-world matrices, which are often approximately low rank with spectra ranging from a few dominant factors to long tails~\citep{udell2019why}. The main training phase mixes four equally likely families: low-rank matrices~\citep{candes2009exact}, two broader-rank spectral families with slowly and steeply decaying singular values (\cref{eq:spectral-prior,eq:prior-spectrum}), and latent-factor matrices~\citep{salakhutdinov2008pmf} (\cref{eq:latent-prior}). Within each family we randomize the rank, the shape of the spectrum, the geometry of the singular vectors and the matrix dimensions (\cref{tab:prior-spectrum-parameters}). Entries are then hidden either uniformly at random or by a mixed scheme whose missingness can depend on the values themselves or follow structured patterns, at rates from 5\% to 75\%; most training examples additionally thin some rows or columns to mimic units with few observations. Later stages add sparse synthetic user--item ratings (\cref{sec:ratings-training}). \Cref{sec:priors-training-full} specifies the families, the masks and the normalization, and \cref{fig:synthetic-priors} shows example masks and spectra.

\section{Results}

\begin{figure}[!htbp]
\centering
\includegraphics[width=0.95\linewidth]{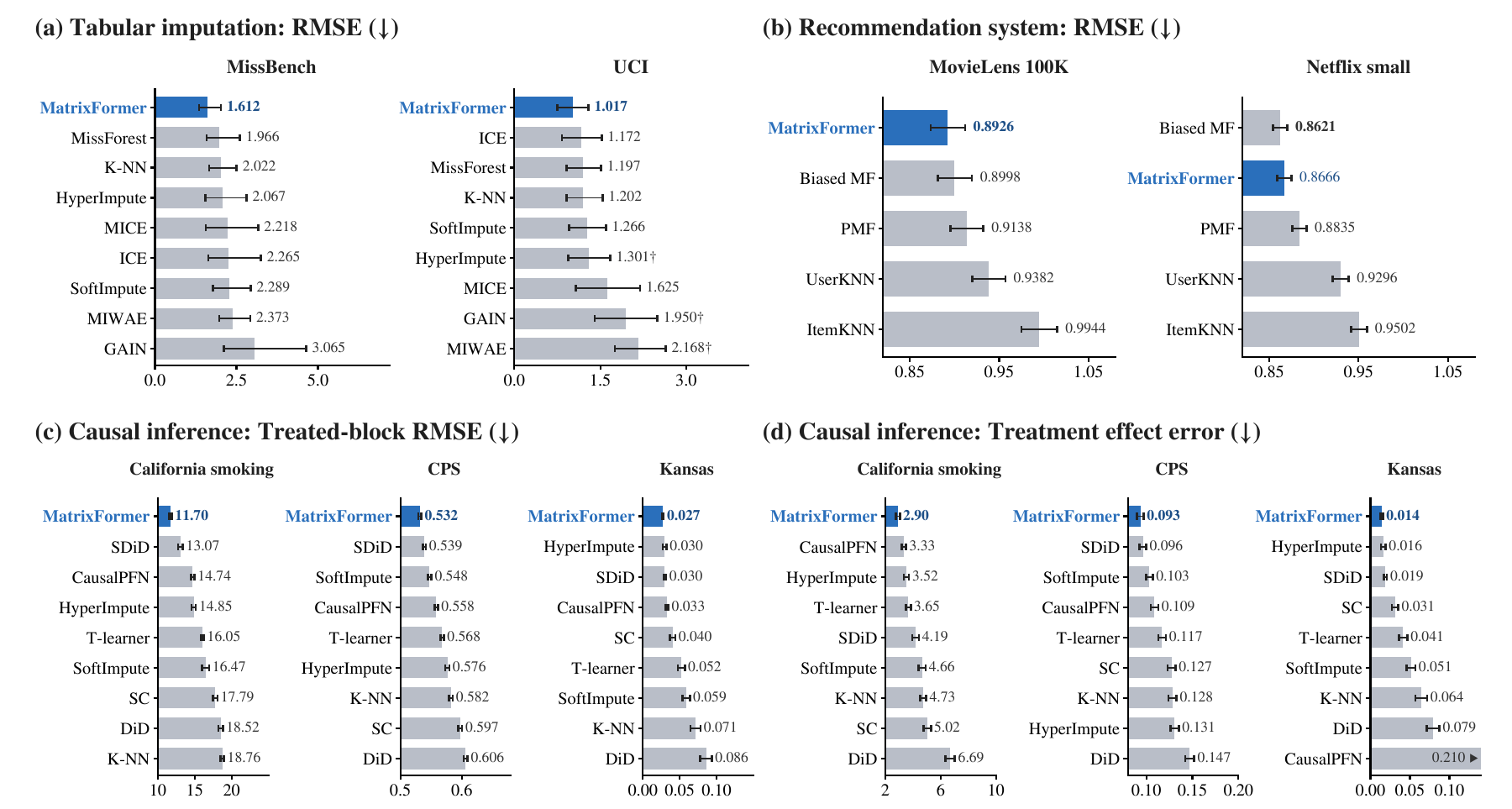}
\caption{\textbf{Comparison of MatrixFormer with baselines on tabular imputation, recommendation and causal panel benchmarks.} MatrixFormer uses one set of weights throughout. Lower is better in every panel; methods are ordered by error and the lowest value is in bold. Error bars show $95\%$ confidence intervals. (a) Imputation error on the MissBench and UCI suites.
(b) Rating-prediction error on MovieLens 100K and Netflix small.
(c) Counterfactual-prediction error on the California smoking, CPS and Kansas placebo benchmarks.
(d) Treatment-effect error on the same benchmarks: the absolute difference $|\hat\tau-\tau|$ between the estimated and the true average effect on the treated units, where the true effect $\tau$ is zero in a placebo.}
\label{fig:empirical-overview}
\end{figure}

We evaluate our pretrained MatrixFormer model across four matrix-completion domains, without ever updating its network weights. \Cref{fig:empirical-overview} summarizes three of the four empirical applications using their existing metrics and evaluation protocols. The fourth, language-model benchmark-score completion, is reported in \cref{sec:llm-appendix}.

\subsection{Tabular imputation}
\label{sec:imputation-results-restructured}
\label{sec:imputation-results}

We evaluate imputation on MissBench, which is a suite of $42$ OpenML datasets under $11$ missingness mechanisms \citep{feitelberg2025tabimpute}, and on our $12$-dataset HyperImpute-inspired UCI suite \citep{jarrett2022hyperimpute}.
Using observed entries as context, MatrixFormer completes whole matrices on MissBench and UCI, with performance measured by columnwise standard-deviation-normalized RMSE averaged across datasets. MatrixFormer achieves overall errors of $1.612$ and $1.017$, respectively, outperforming the eight shared baselines on both suites. Moreover, the remaining methods are each fit on each such matrix and then used for imputation, whereas, our results are simply via inference and in context learning. See \cref{fig:empirical-overview}(a) for the results and
Appendix~\ref{sec:tabular-appendix} for extensive benchmark and evaluation details.

\subsection{Matrix completion for recommendation}
\label{sec:recsys-results-restructured}
\label{sec:recsys-results}

We evaluate rating prediction on MovieLens $100$K \citep{harper2015movielens} and the Netflix small subset \citep{salah2020cornac},
which are two sparse user movie matrices containing ratings on a $1$--$5$ scale. Using $80\%$ of the available ratings as context, MatrixFormer predicts the held-out $20\%$ without updating its pretrained network weights. We compute RMSE separately for each user and then average across users. MatrixFormer achieves the lowest RMSE among the displayed methods on MovieLens $100$K ($0.8926$), while on Netflix small its RMSE of $0.8666$ is slightly higher than Biased MF's $0.8621$ and lower than the remaining baselines.\footnote{We note that most of the top methods in this domain are within error bars and the result should be viewed as MatrixFormer ties the best previous algorithms.} See \cref{fig:empirical-overview}(b) for the results and
Appendix~\ref{sec:recsys-appendix} for further details.

\subsection{Causal inference}
\label{sec:causal-results}

Estimating a treatment effect from panel data is a matrix completion problem \citep{ding2018causal,athey2021matrix}: in a unit-by-time matrix of outcomes, the entries of treated units after treatment are missing counterfactuals, namely the outcomes those units would have had without treatment. We evaluate MatrixFormer on three placebo benchmarks built from California cigarette-sales data \citep{abadie2010synthetic}, CPS labor-market data \citep{arkhangelsky2021synthetic} and the Kansas semi-synthetic design \citep{benmichael2021augmented}.
MatrixFormer predicts these missing counterfactuals from the observed entries in one forward pass. We report the RMSE of the counterfactual predictions and the error of the treatment effect they imply, whose true value is zero. MatrixFormer achieves the lowest RMSE and the lowest treatment effect error on all three benchmarks, ahead of both synthetic difference-in-differences (SDiD) \citep{arkhangelsky2021synthetic}, a method designed for panel data, and CausalPFN \citep{balazadeh2025causalpfn}, a causal foundation model. See \cref{fig:empirical-overview}(c) and (d) for the results and Appendix~\ref{sec:causal-protocol} for further details.

\section{Interpretability}
\label{sec:interpretability}

Here, we demonstrate that MatrixFormer encodes a representation of latent factors for rows and columns and is able to recover inherent groupings in matrices. Note that these behaviors were \emph{emergent}: we did not explicitly train MatrixFormer for these behaviors.

\subsection{Latent factors in the CLS tokens}

\label{sec:cls-factors}
\begin{figure}[H]
\centering
\includegraphics[width=\linewidth]{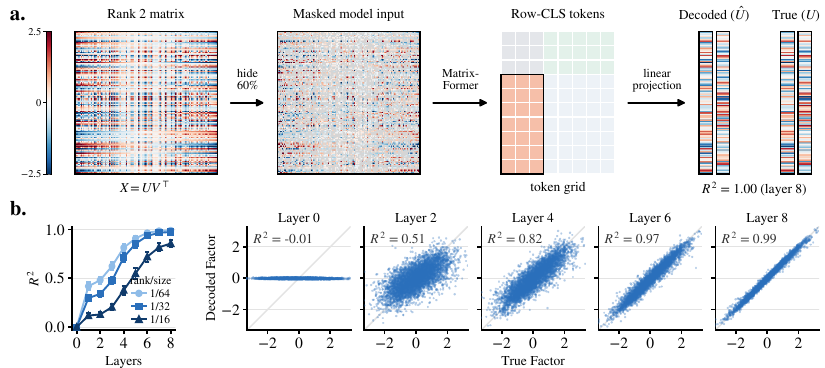}
\caption{\textbf{CLS tokens recover latent factors.} (a) We create a rank-2 matrix and hide 60\% of its entries. A linear projection of the row-CLS tokens gives the decoded row factors $\widehat U$. (b) As the tokens go through attention layers, the decoded factors align with the true factors.
}
\label{fig:cls-factors}
\end{figure}

From a partially observed low-rank matrix, the CLS tokens of MatrixFormer recover the underlying latent factors up to a linear projection, purely in context: no weights are updated, and the factors are never given as input. We test this on matrices $X=UV^\top$ of size $n$ and rank $r$ with standard-Gaussian factors and $40\%$ of the entries observed uniformly at random; we consider rank-to-size ratios $r/n$ of $1/64$, $1/32$ and $1/16$ (\cref{fig:cls-factors}a shows one such matrix). Write $U_i\in\mathbb{R}^r$ for the $i$-th row of $U$, the row factor of row $i$. At each layer $\ell$, we concatenate the 12 row-CLS tokens of row $i$ into one vector $c_i(\ell)$ and fit a linear projection from $c_i(\ell)$ to $U_i$ by ridge regression on seven eighths of the rows, computing $R^2$ only on the held-out rows, over eight folds.
Averaged over the factor coordinates and over 30 matrices, this held-out $R^2$ increases from zero before attention, where the tokens are identical across rows, to $0.99$ after the last block of attention at ratio $1/64$ (\cref{fig:cls-factors}b); it reaches $0.97$ and $0.86$ at ratios $1/32$ and $1/16$. The scatter plots pool the projected and true factor coordinates of three matrices, of sizes 128, 256 and 512. The column-CLS tokens recover the column factors to the same degree (\cref{fig:cls-factors-columns}); the full setup is in \cref{sec:cls-probe-details}.

\subsection{Recovery of latent group structure}
\label{sec:synthetic-results}

\begin{figure}[!htbp]
\centering
\includegraphics[width=\linewidth]{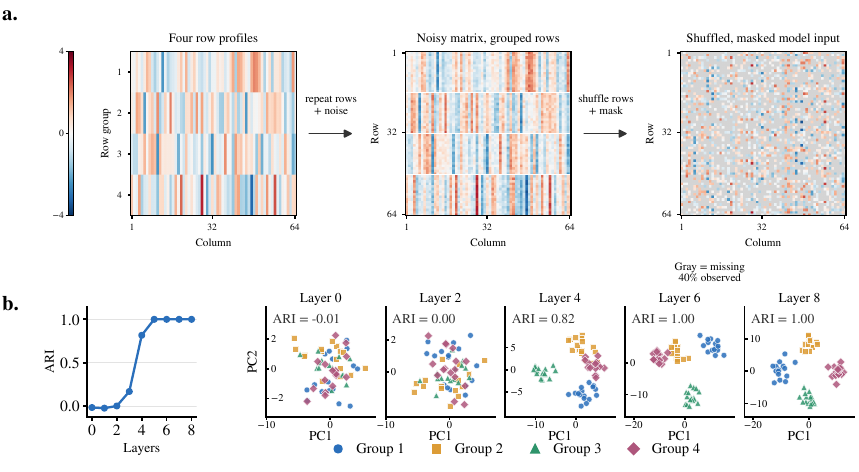}
\caption{\textbf{Embeddings recover latent groups.} (a) We create a matrix with 4 row-wise groups, Gaussian noise, and shuffled entries. (b) As the embeddings go through attention layers, these 4 groups become clustered. A simple K-Means with $K=4$ recovers the groups after layer 5.}
\label{fig:synthetic-embedding-example}
\end{figure}

We illustrate latent-group recovery using a single 64-by-64 matrix with four row-wise groups, Gaussian noise with standard deviation 0.25, and 40\% observed entries under MCAR masking. MatrixFormer receives only observed values and the mask, not group labels. At each layer, we average cell embeddings within each row and apply K-means with four clusters to the full 768-dimensional row embeddings. Adjusted Rand index (ARI) measures agreement with the true row groups, increasing from -0.014 before attention to 1.000 after block eight as shown in \cref{fig:synthetic-embedding-example}. The scatter plots use PCA only for visualization. We can see clearly that the 4 row-wise groups are easily recoverable from the embeddings alone.

\section{Conclusion}

We introduced MatrixFormer, a model for matrix completion that represents a partially observed matrix as a grid of cell tokens, alternates attention over rows and columns, and is pretrained only on synthetic matrices. With no fine-tuning, the same mixed-prior network was applied to tabular imputation, rating prediction, benchmark-score completion for language models and causal panel data. MatrixFormer has lower overall RMSE than the displayed baselines on UCI, while TabImpute is slightly better on MissBench. It improves on the collaborative-filtering baselines on MovieLens 100K, but Biased MF has lower error on Netflix small. At MI budget $k=15$, MatrixFormer has higher $R^2$ than kernel ridge on Merged and BenchPress, but lower $R^2$ on MMLU and MTEB. On causal panels, MatrixFormer has the lowest normalized point estimates among the displayed methods for both metrics on all three benchmarks. The main comparisons retain standard orientation across domains. Transposition and prediction averaging are reported as exploratory ablations, with improvements and regressions retained rather than used to select a primary result. A single-network instance of the MatrixFormer architecture completes a $1024\times10$ matrix $107\times$ faster than TabImpute; the orientation-averaging ablation uses two passes of the single model. Evaluating the predictive distributions, for example as intervals for treatment effects, and extending the context to larger matrices are directions for future work. These results show cross-domain transfer without implying uniform superiority over task-specific methods.

\bibliographystyle{plainnat}
\bibliography{references}

@misc{smola2026submodular,
  author = {Smola, Alexander},
  title = {Submodular Benchmark Selection},
  year = {2026},
  eprint = {2605.02209},
  archivePrefix = {arXiv},
  primaryClass = {cs.AI},
  doi = {10.48550/arXiv.2605.02209},
  url = {https://arxiv.org/abs/2605.02209},
  note = {Version 1; accompanying code and data release}
}

@inproceedings{muennighoff2023mteb,
  author = {Muennighoff, Niklas and Tazi, Nouamane and Magne, Loic and Reimers, Nils},
  title = {{MTEB}: Massive Text Embedding Benchmark},
  booktitle = {Proceedings of the 17th Conference of the European Chapter of the Association for Computational Linguistics},
  year = {2023},
  pages = {2014--2037},
  publisher = {Association for Computational Linguistics},
  doi = {10.18653/v1/2023.eacl-main.148},
  url = {https://aclanthology.org/2023.eacl-main.148/}
}

@inproceedings{hendrycks2021mmlu,
  author = {Hendrycks, Dan and Burns, Collin and Basart, Steven and Zou, Andy and Mazeika, Mantas and Song, Dawn and Steinhardt, Jacob},
  title = {Measuring Massive Multitask Language Understanding},
  booktitle = {International Conference on Learning Representations},
  year = {2021},
  url = {https://openreview.net/forum?id=d7KBjmI3GmQ}
}

@misc{benchpress2026matrix,
  author = {{BenchPress contributors}},
  title = {{LLM} Benchmark Matrix Completion},
  year = {2026},
  howpublished = {GitHub repository},
  url = {https://github.com/anadim/llm-benchmark-matrix},
  note = {Includes the BenchPress method; evaluated score snapshot generated February 24, 2026, as distributed with Submodular Benchmark Selection}
}

@misc{feitelberg2025tabimpute,
  author = {Feitelberg, Jacob and Saha, Dwaipayan and Choi, Kyuseong and Ahmad, Zaid and Agarwal, Anish and Dwivedi, Raaz},
  title = {One Pipeline, Many Transformers: Pattern-Specific Imputation Specialists for Tabular Missing Data},
  year = {2025},
  eprint = {2510.02625},
  archivePrefix = {arXiv},
  primaryClass = {cs.LG},
  doi = {10.48550/arXiv.2510.02625},
  url = {https://arxiv.org/abs/2510.02625v6},
  note = {Version 6, revised August 2026; first submitted October 2025}
}

@inproceedings{jarrett2022hyperimpute,
  author = {Jarrett, Daniel and Cebere, Bogdan C and Liu, Tennison and Curth, Alicia and van der Schaar, Mihaela},
  title = {{HyperImpute}: Generalized Iterative Imputation with Automatic Model Selection},
  booktitle = {Proceedings of the 39th International Conference on Machine Learning},
  series = {Proceedings of Machine Learning Research},
  volume = {162},
  pages = {9916--9937},
  year = {2022},
  publisher = {PMLR},
  url = {https://proceedings.mlr.press/v162/jarrett22a.html}
}

@article{athey2021matrix,
  title = {Matrix Completion Methods for Causal Panel Data Models},
  author = {Athey, Susan and Bayati, Mohsen and Doudchenko, Nikolay and Imbens, Guido and Khosravi, Khashayar},
  journal = {Journal of the American Statistical Association},
  volume = {116},
  number = {536},
  pages = {1716--1730},
  year = {2021}
}

@article{arkhangelsky2021synthetic,
  title = {Synthetic Difference-in-Differences},
  author = {Arkhangelsky, Dmitry and Athey, Susan and Hirshberg, David A. and Imbens, Guido W. and Wager, Stefan},
  journal = {American Economic Review},
  volume = {111},
  number = {12},
  pages = {4088--4118},
  year = {2021}
}

@article{benmichael2021augmented,
  title = {The Augmented Synthetic Control Method},
  author = {Ben-Michael, Eli and Feller, Avi and Rothstein, Jesse},
  journal = {Journal of the American Statistical Association},
  volume = {116},
  number = {536},
  pages = {1789--1803},
  year = {2021}
}

@article{abadie2010synthetic,
  title = {Synthetic Control Methods for Comparative Case Studies: Estimating the Effect of {California's} Tobacco Control Program},
  author = {Abadie, Alberto and Diamond, Alexis and Hainmueller, Jens},
  journal = {Journal of the American Statistical Association},
  volume = {105},
  number = {490},
  pages = {493--505},
  year = {2010}
}

@inproceedings{gorla2025cacti,
  author = {Gorla, Aditya and Wang, Ryan and Liu, Zhengtong and An, Ulzee and Sankararaman, Sriram},
  title = {{CACTI}: Leveraging Copy Masking and Contextual Information to Improve Tabular Data Imputation},
  booktitle = {Proceedings of the 42nd International Conference on Machine Learning},
  year = {2025},
  url = {https://arxiv.org/abs/2506.02306}
}

@article{harper2015movielens,
  title = {The {MovieLens} Datasets: History and Context},
  author = {Harper, F. Maxwell and Konstan, Joseph A.},
  journal = {ACM Transactions on Interactive Intelligent Systems},
  volume = {5},
  number = {4},
  pages = {1--19},
  year = {2015},
  doi = {10.1145/2827872}
}

@article{salah2020cornac,
  title = {{Cornac}: A Comparative Framework for Multimodal Recommender Systems},
  author = {Salah, Aghiles and Truong, Quoc-Tuan and Lauw, Hady W.},
  journal = {Journal of Machine Learning Research},
  volume = {21},
  number = {95},
  pages = {1--5},
  year = {2020},
  url = {https://jmlr.org/papers/v21/19-805.html}
}

@inproceedings{bennett2007netflix,
  title={The netflix prize},
  author={Bennett, James and Lanning, Stan and others},
  booktitle={Proceedings of KDD cup and workshop},
  volume={2007},
  pages={35},
  year={2007},
  organization={New York}
}

@article{koren2009bellkor,
  title={The bellkor solution to the netflix grand prize},
  author={Koren, Yehuda},
  journal={Netflix prize documentation},
  volume={81},
  number={2009},
  pages={1--10},
  year={2009}
}

@article{vanbuuren2011mice,
  title = {{mice}: Multivariate Imputation by Chained Equations in {R}},
  author = {van Buuren, Stef and Groothuis-Oudshoorn, Karin},
  journal = {Journal of Statistical Software},
  volume = {45},
  number = {3},
  pages = {1--67},
  year = {2011},
  doi = {10.18637/jss.v045.i03}
}

@inproceedings{vaswani2017attention,
  author = {Vaswani, Ashish and Shazeer, Noam and Parmar, Niki and Uszkoreit, Jakob and Jones, Llion and Gomez, Aidan N. and Kaiser, Lukasz and Polosukhin, Illia},
  title = {Attention Is All You Need},
  booktitle = {Advances in Neural Information Processing Systems},
  volume = {30},
  year = {2017},
  url = {https://proceedings.neurips.cc/paper/2017/hash/3f5ee243547dee91fbd053c1c4a845aa-Abstract.html}
}

@inproceedings{devlin2019bert,
  author = {Devlin, Jacob and Chang, Ming-Wei and Lee, Kenton and Toutanova, Kristina},
  title = {{BERT}: Pre-training of Deep Bidirectional Transformers for Language Understanding},
  booktitle = {Proceedings of the 2019 Conference of the North American Chapter of the Association for Computational Linguistics: Human Language Technologies, Volume 1 (Long and Short Papers)},
  pages = {4171--4186},
  year = {2019},
  publisher = {Association for Computational Linguistics},
  doi = {10.18653/v1/N19-1423},
  url = {https://aclanthology.org/N19-1423/}
}

@misc{ho2019axial,
  author = {Ho, Jonathan and Kalchbrenner, Nal and Weissenborn, Dirk and Salimans, Tim},
  title = {Axial Attention in Multidimensional Transformers},
  year = {2019},
  eprint = {1912.12180},
  archivePrefix = {arXiv},
  primaryClass = {cs.CV},
  doi = {10.48550/arXiv.1912.12180},
  url = {https://arxiv.org/abs/1912.12180}
}

@inproceedings{xiong2020layernorm,
  author = {Xiong, Ruibin and Yang, Yunchang and He, Di and Zheng, Kai and Zheng, Shuxin and Xing, Chen and Zhang, Huishuai and Lan, Yanyan and Wang, Liwei and Liu, Tieyan},
  title = {On Layer Normalization in the Transformer Architecture},
  booktitle = {Proceedings of the 37th International Conference on Machine Learning},
  series = {Proceedings of Machine Learning Research},
  volume = {119},
  pages = {10524--10533},
  year = {2020},
  publisher = {PMLR},
  url = {https://proceedings.mlr.press/v119/xiong20b.html}
}

@misc{su2021roformer,
  author = {Su, Jianlin and Lu, Yu and Pan, Shengfeng and Murtadha, Ahmed and Wen, Bo and Liu, Yunfeng},
  title = {{RoFormer}: Enhanced Transformer with Rotary Position Embedding},
  year = {2021},
  eprint = {2104.09864},
  archivePrefix = {arXiv},
  primaryClass = {cs.CL},
  doi = {10.48550/arXiv.2104.09864},
  url = {https://arxiv.org/abs/2104.09864}
}

@misc{shazeer2020glu,
  author = {Shazeer, Noam},
  title = {{GLU} Variants Improve Transformer},
  year = {2020},
  eprint = {2002.05202},
  archivePrefix = {arXiv},
  primaryClass = {cs.LG},
  doi = {10.48550/arXiv.2002.05202},
  url = {https://arxiv.org/abs/2002.05202}
}

@article{hollmann2025tabpfn,
  author = {Hollmann, Noah and M{\"u}ller, Samuel and Purucker, Lennart and Krishnakumar, Arjun and K{\"o}rfer, Max and Hoo, Shi Bin and Schirrmeister, Robin Tibor and Hutter, Frank},
  title = {Accurate Predictions on Small Data with a Tabular Foundation Model},
  journal = {Nature},
  volume = {637},
  pages = {319--326},
  year = {2025}
}

@inproceedings{qu2025tabicl,
  author = {Qu, Jingang and Holzm{\"u}ller, David and Varoquaux, Ga{\"e}l and Le Morvan, Marine},
  title = {{TabICL}: A Tabular Foundation Model for In-Context Learning on Large Data},
  booktitle = {Proceedings of the 42nd International Conference on Machine Learning},
  year = {2025}
}

@inproceedings{zhang2019rmsnorm,
  author = {Zhang, Biao and Sennrich, Rico},
  title = {Root Mean Square Layer Normalization},
  booktitle = {Advances in Neural Information Processing Systems},
  volume = {32},
  year = {2019}
}

@inproceedings{bachlechner2021rezero,
  author = {Bachlechner, Thomas and Majumder, Bodhisattwa Prasad and Mao, Huanru Henry and Cottrell, Garrison W. and McAuley, Julian},
  title = {{ReZero} is All You Need: Fast Convergence at Large Depth},
  booktitle = {Proceedings of the Thirty-Seventh Conference on Uncertainty in Artificial Intelligence},
  series = {Proceedings of Machine Learning Research},
  volume = {161},
  year = {2021}
}

@inproceedings{muller2022pfn,
  author = {M{\"u}ller, Samuel and Hollmann, Noah and Pineda Arango, Sebastian and Grabocka, Josif and Hutter, Frank},
  title = {Transformers Can Do {B}ayesian Inference},
  booktitle = {International Conference on Learning Representations},
  year = {2022}
}

@article{epstein1969rps,
  author = {Epstein, Edward S.},
  title = {A Scoring System for Probability Forecasts of Ranked Categories},
  journal = {Journal of Applied Meteorology},
  volume = {8},
  number = {6},
  pages = {985--987},
  year = {1969}
}

@article{gneiting2007scoring,
  author = {Gneiting, Tilmann and Raftery, Adrian E.},
  title = {Strictly Proper Scoring Rules, Prediction, and Estimation},
  journal = {Journal of the American Statistical Association},
  volume = {102},
  number = {477},
  pages = {359--378},
  year = {2007}
}

@article{huber1964robust,
  author = {Huber, Peter J.},
  title = {Robust Estimation of a Location Parameter},
  journal = {The Annals of Mathematical Statistics},
  volume = {35},
  number = {1},
  pages = {73--101},
  year = {1964}
}

@article{hinton2015distilling,
  author = {Hinton, Geoffrey and Vinyals, Oriol and Dean, Jeff},
  title = {Distilling the Knowledge in a Neural Network},
  journal = {arXiv preprint arXiv:1503.02531},
  year = {2015}
}

@inproceedings{loshchilov2019adamw,
  author = {Loshchilov, Ilya and Hutter, Frank},
  title = {Decoupled Weight Decay Regularization},
  booktitle = {International Conference on Learning Representations},
  year = {2019}
}

@inproceedings{loshchilov2017sgdr,
  author = {Loshchilov, Ilya and Hutter, Frank},
  title = {{SGDR}: Stochastic Gradient Descent with Warm Restarts},
  booktitle = {International Conference on Learning Representations},
  year = {2017}
}

@article{candes2009exact,
  author = {Cand{\`e}s, Emmanuel J. and Recht, Benjamin},
  title = {Exact Matrix Completion via Convex Optimization},
  journal = {Foundations of Computational Mathematics},
  volume = {9},
  number = {6},
  pages = {717--772},
  year = {2009}
}

@article{udell2019why,
  author = {Udell, Madeleine and Townsend, Alex},
  title = {Why Are Big Data Matrices Approximately Low Rank?},
  journal = {SIAM Journal on Mathematics of Data Science},
  volume = {1},
  number = {1},
  pages = {144--160},
  year = {2019}
}

@inproceedings{salakhutdinov2008pmf,
  author = {Salakhutdinov, Ruslan and Mnih, Andriy},
  title = {Probabilistic Matrix Factorization},
  booktitle = {Advances in Neural Information Processing Systems},
  volume = {20},
  year = {2007},
  url = {https://proceedings.neurips.cc/paper_files/paper/2007/hash/d7322ed717dedf1eb4e6e52a37ea7bcd-Abstract.html}
}

@article{rubin1976inference,
  author = {Rubin, Donald B.},
  title = {Inference and Missing Data},
  journal = {Biometrika},
  volume = {63},
  number = {3},
  pages = {581--592},
  year = {1976}
}

@article{robins1995catastrophic,
  author = {Robins, Anthony},
  title = {Catastrophic Forgetting, Rehearsal and Pseudorehearsal},
  journal = {Connection Science},
  volume = {7},
  number = {2},
  pages = {123--146},
  year = {1995}
}

@article{koren2009matrix,
  author = {Koren, Yehuda and Bell, Robert and Volinsky, Chris},
  title = {Matrix Factorization Techniques for Recommender Systems},
  journal = {Computer},
  volume = {42},
  number = {8},
  pages = {30--37},
  year = {2009}
}

@inproceedings{resnick1994grouplens,
  author = {Resnick, Paul and Iacovou, Neophytos and Suchak, Mitesh and Bergstrom, Peter and Riedl, John},
  title = {{GroupLens}: An Open Architecture for Collaborative Filtering of Netnews},
  booktitle = {Proceedings of the 1994 ACM Conference on Computer Supported Cooperative Work},
  pages = {175--186},
  year = {1994}
}

@inproceedings{sarwar2001item,
  author = {Sarwar, Badrul and Karypis, George and Konstan, Joseph and Riedl, John},
  title = {Item-Based Collaborative Filtering Recommendation Algorithms},
  booktitle = {Proceedings of the 10th International Conference on World Wide Web},
  pages = {285--295},
  year = {2001}
}

@article{troyanskaya2001missing,
  author = {Troyanskaya, Olga and Cantor, Michael and Sherlock, Gavin and Brown, Pat and Hastie, Trevor and Tibshirani, Robert and Botstein, David and Altman, Russ B.},
  title = {Missing Value Estimation Methods for {DNA} Microarrays},
  journal = {Bioinformatics},
  volume = {17},
  number = {6},
  pages = {520--525},
  year = {2001}
}

@article{mazumder2010spectral,
  author = {Mazumder, Rahul and Hastie, Trevor and Tibshirani, Robert},
  title = {Spectral Regularization Algorithms for Learning Large Incomplete Matrices},
  journal = {Journal of Machine Learning Research},
  volume = {11},
  pages = {2287--2322},
  year = {2010}
}

@article{stekhoven2012missforest,
  author = {Stekhoven, Daniel J. and B{\"u}hlmann, Peter},
  title = {{MissForest}---Non-Parametric Missing Value Imputation for Mixed-Type Data},
  journal = {Bioinformatics},
  volume = {28},
  number = {1},
  pages = {112--118},
  year = {2012}
}

@inproceedings{yoon2018gain,
  author = {Yoon, Jinsung and Jordon, James and van der Schaar, Mihaela},
  title = {{GAIN}: Missing Data Imputation using Generative Adversarial Nets},
  booktitle = {Proceedings of the 35th International Conference on Machine Learning},
  pages = {5689--5698},
  year = {2018}
}

@inproceedings{mattei2019miwae,
  author = {Mattei, Pierre-Alexandre and Frellsen, Jes},
  title = {{MIWAE}: Deep Generative Modelling and Imputation of Incomplete Data Sets},
  booktitle = {Proceedings of the 36th International Conference on Machine Learning},
  pages = {4413--4423},
  year = {2019}
}

@article{balazadeh2025causalpfn,
  title = {{CausalPFN}: Amortized Causal Effect Estimation via In-Context Learning},
  author = {Balazadeh, Vahid and Kamkari, Hamidreza and Thomas, Valentin and Li, Benson and Ma, Junwei and Cresswell, Jesse C. and Krishnan, Rahul G.},
  journal = {arXiv preprint arXiv:2506.07918},
  year = {2025}
}

@article{kunzel2019metalearners,
  title = {Metalearners for Estimating Heterogeneous Treatment Effects using Machine Learning},
  author = {K{\"u}nzel, S{\"o}ren R. and Sekhon, Jasjeet S. and Bickel, Peter J. and Yu, Bin},
  journal = {Proceedings of the National Academy of Sciences},
  volume = {116},
  number = {10},
  pages = {4156--4165},
  year = {2019}
}

@article{ding2018causal,
  title = {Causal Inference: A Missing Data Perspective},
  author = {Ding, Peng and Li, Fan},
  journal = {Statistical Science},
  volume = {33},
  number = {2},
  pages = {214--237},
  year = {2018}
}

\appendix

\section{Model architecture}
\label{sec:architecture-full}

\paragraph{Cell representation.}
We form $\widetilde X\in\mathbb{R}^{R\times C}$ by setting $\widetilde X_{rc}=X_{rc}$ at observed entries and $\widetilde X_{rc}=0$ at missing entries. Zero serves only as a computational placeholder: the corresponding value embedding is subsequently replaced by a learned mask token, as in masked-token modeling~\citep{devlin2019bert}. Let $d$ be the embedding dimension, let $f_{\mathrm{val}}:\mathbb{R}\to\mathbb{R}^{d}$ be a shared value-embedding multilayer perceptron (MLP), let $m_{\mathrm{mask}}\in\mathbb{R}^{d}$ be the learned mask vector, and let $e_{\mathrm{miss}}:\{0,1\}\to\mathbb{R}^{d}$ be an embedding lookup with separate learned vectors for observed and missing cells. The initial representation of cell $(r,c)$ is
\begin{equation}
E_{rc}^{(0)}
=
(1-M_{rc}) f_{\mathrm{val}}(\widetilde X_{rc})
+M_{rc}m_{\mathrm{mask}}
+e_{\mathrm{miss}}(M_{rc}).
\label{eq:cell-embedding}
\end{equation}
Thus, observed cells receive $f_{\mathrm{val}}(X_{rc})+e_{\mathrm{miss}}(0)$, whereas missing cells receive $m_{\mathrm{mask}}+e_{\mathrm{miss}}(1)$. The mask token already encodes missingness; the final term provides a separately parameterized offset for each observation status.

\paragraph{Alternating axial attention.}
Each transformer block contains two axial self-attention operations~\citep{ho2019axial}. \emph{Feature attention} is applied across columns separately within each row, allowing a cell to use the other variables measured for the same datapoint. \emph{Datapoint attention} is then applied across rows separately within each column, allowing the same cell to use the empirical structure of that variable across observations. We write $\operatorname{RMSNorm}$ for root-mean-square normalization over the $d$ embedding coordinates of each token~\citep{zhang2019rmsnorm}, and $\operatorname{FFN}:\mathbb{R}^{d}\to\mathbb{R}^{d}$ for a feed-forward network with SwiGLU gating~\citep{shazeer2020glu}, applied separately at each grid location with shared parameters. Both attention operations and the feed-forward sublayer use pre-normalized residual connections~\citep{xiong2020layernorm}, each scaled by a learned scalar gain $\gamma_i$~\citep{bachlechner2021rezero}. Writing $\mathcal{A}_{\mathrm{feat}}$ and $\mathcal{A}_{\mathrm{data}}$ for the two attention operators, one block can be summarized as
\begin{align}
U^{(\ell)} &= G^{(\ell-1)} + \gamma_1\,\mathcal{A}_{\mathrm{feat}}\!\left(\operatorname{RMSNorm}(G^{(\ell-1)})\right),\\
V^{(\ell)} &= U^{(\ell)} + \gamma_2\,\mathcal{A}_{\mathrm{data}}\!\left(\operatorname{RMSNorm}(U^{(\ell)})\right),\\
G^{(\ell)} &= V^{(\ell)} + \gamma_3\,\operatorname{FFN}\!\left(\operatorname{RMSNorm}(V^{(\ell)})\right),
\end{align}
for $\ell=1,\ldots,L$, with layer- and sublayer-specific parameters suppressed in the notation. Rotary positional embeddings (RoPE)~\citep{su2021roformer} are applied along the attended axis, using column positions for feature attention and row positions for datapoint attention, with data positions beginning after the corresponding CLS border. Alternating the two attention operators lets information move across the full matrix while retaining the distinction between row structure and column structure at every layer. The feed-forward network mixes embedding coordinates rather than matrix entries. We use $L=8$ blocks of width $d=768$ with 16 attention heads, feed-forward width 1536 and $K_r=K_c=12$, for 78M parameters (Table~\ref{tab:arch-hparams}).

\paragraph{Decoder.}
After the final transformer block, we remove the row- and column-CLS border and retain the contextual representation $h_{rc}=G^{(L)}_{K_r+r,K_c+c}\in\mathbb{R}^{d}$ for each matrix entry. Let $\theta$ collect the trainable model parameters and let $f_{\mathrm{dec}}:\mathbb{R}^{d}\to\mathbb{R}^{O}$ be the shared decoder MLP. Its output $z\in\mathbb{R}^{O}$ parameterizes a bar distribution~\citep{muller2022pfn,hollmann2025tabpfn} on $O=5000$ fixed bins with borders $b_0<b_1<\dots<b_O$ and midpoints $\tilde b_k=(b_{k-1}+b_k)/2$: with bin probabilities $q(z)=\mathrm{softmax}(z)$, its density $p(y;z)$ at a scalar target $y$ equals $q_k(z)/(b_k-b_{k-1})$ on each interior bin $[b_{k-1},b_k)$, and the two outer bins are replaced by half-normal tails so that the support is $\mathbb{R}$ (Appendix~\ref{sec:bar-distribution}). The distribution has no parameters of its own, so $\theta$ enters only through $z$. The conditional predictive density for cell $(r,c)$ is then
\begin{equation}
\begin{aligned}
 z_{rc} &= f_{\mathrm{dec}}(h_{rc})\in\mathbb{R}^{O},\\
 p_{\theta,rc}(y\mid X,M) &= p(y;z_{rc}).
\end{aligned}
 \label{eq:cell-decoder}
\end{equation}
The dependence of $h_{rc}$ and $f_{\mathrm{dec}}$ on $\theta$ is implicit. The decoder acts separately on each token, with parameters shared across rows and columns; each prediction nevertheless uses matrix-wide context through $h_{rc}$, rather than through an additional mixing operation in the output head. This cellwise readout permits all missing entries to be predicted in parallel within a single forward pass. For point imputation, we use the predictive mean,
\begin{equation}
 \widehat X_{rc}
 =\mu_{\theta,rc}
 =\mathbb{E}_{Y\sim p_{\theta,rc}(\,\cdot\mid X,M)}[Y]
 =\sum_{k=1}^{O}q_k(z_{rc})\,\nu_k,
 \qquad M_{rc}=1,
 \label{eq:decoder-mean}
\end{equation}
where $\nu_k=\tilde b_k$ for $1<k<O$ and $\nu_1,\nu_O$ are the means of the two tails; observed entries are left unchanged. Retaining a predictive distribution, rather than only a scalar output, allows the same decoder to support the likelihood and cumulative-distribution-function (CDF) alignment terms in the training objective, while its mean supplies the point-prediction term. During synthetic training, these losses are evaluated only at entries hidden by the training mask, as described in \cref{sec:priors-training-full}. When discussing a single hidden entry, we suppress $(r,c)$ and write $p_\theta$ for $p_{\theta,rc}$.

\paragraph{Computational scaling.}
Holding the embedding width fixed, dense self-attention applied to the flattened $RC$ cells requires $\mathcal{O}(R^2C^2)$ pairwise attention interactions per layer. Ignoring the CLS borders, MatrixFormer instead performs attention over $C$ tokens for each of $R$ rows and over $R$ tokens for each of $C$ columns, giving
\begin{equation}
\mathcal{O}\!\left(RC^2+CR^2\right)
=\mathcal{O}\!\left(RC(R+C)\right).
\end{equation}
Including the border tokens replaces $R$ and $C$ by $R+K_r$ and $C+K_c$, respectively, leaving the asymptotic order unchanged for fixed $K_r$ and $K_c$. Compared with full attention on the flattened grid, this factorization reduces the attention cost while preserving the row--column structure of the problem.

\subsection{Architecture}
\label{sec:architecture-hparams}
\Cref{tab:arch-hparams} lists the settings of the network in \cref{sec:architecture}.

\begin{table}[ht]
\centering
\caption{MatrixFormer architecture.}
\label{tab:arch-hparams}
\small
\begin{tabular}{@{}lp{0.66\linewidth}@{}}
\toprule
Component & Setting \\
\midrule
Value embedding $f_{\mathrm{val}}$ & MLP $1\to768\to768$ with GELU \\
Missing entries & learned mask vector $m_{\mathrm{mask}}$; learned two-entry missingness embedding $e_{\mathrm{miss}}$ added to every cell \\
CLS border & $K_c=12$ row-CLS tokens $t^{\mathrm{row}}_j$ and $K_r=12$ column-CLS tokens $t^{\mathrm{col}}_i$; corner entry $(i,j)$ is $t^{\mathrm{col}}_i+t^{\mathrm{row}}_j$ \\
Blocks & $L=8$, width $d=768$; feature attention, datapoint attention, feed-forward \\
Attention & 16 heads of dimension 48; rotary embeddings (base $10^4$) on 24 of the 48 dimensions; positions start after the border \\
Feed-forward & SwiGLU, $W_3\big(\mathrm{SiLU}(W_1h)\odot W_2h\big)$, hidden width 1536 \\
Normalization & pre-norm RMSNorm with learned gain, $\epsilon=10^{-6}$; no final normalization \\
Residual gains & one learned scalar per sublayer, initialized to 0.5 \\
Decoder $f_{\mathrm{dec}}$ & MLP $768\to1536\to1536\to5000$ with GELU; $O=5000$ bins \\
Size & 78.0M parameters; all linear maps have biases \\
\bottomrule
\end{tabular}
\end{table}

\subsection{Bar distribution}
\label{sec:bar-distribution}
With the borders $b_k$ and bin probabilities $q(z)$ of \cref{sec:architecture} and widths $w_k=b_k-b_{k-1}$, the full density is
\begin{equation}
p(y;z)=
\begin{cases}
q_1(z)\,\varphi_{s_1}(b_1-y), & y<b_1,\\
q_k(z)/w_k, & b_{k-1}\le y<b_k,\quad 1<k<O,\\
q_O(z)\,\varphi_{s_O}(y-b_{O-1}), & y\ge b_{O-1},
\end{cases}
\label{eq:bar-distribution}
\end{equation}
where $\varphi_s(t)=\sqrt{2/\pi}\,s^{-1}e^{-t^2/(2s^2)}$, $t\ge0$, is the half-normal density and $s_k=w_k/\Phi^{-1}(3/4)$, so that each tail has median equal to its bin width. The family has no trainable parameters; $\theta$ enters only through the logits, $p_{\theta,rc}(\cdot\mid X,M)=p(\cdot\,;z_{rc})$. The tail means in $\mu_{\theta,rc}$ are $\nu_1=b_1-s_1\sqrt{2/\pi}$ and $\nu_O=b_{O-1}+s_O\sqrt{2/\pi}$.

\subsection{Inference runtime scaling}
\label{sec:runtime-results}

\begin{figure}[htbp]
  \centering
  \includegraphics[width=\linewidth]{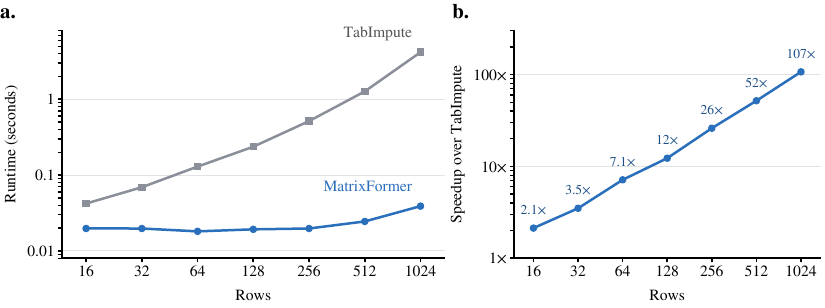}
  \caption{\textbf{Inference runtime of MatrixFormer and TabImpute.} Matrices have 10 columns and $40\%$ of their entries missing completely at random; timings are on one NVIDIA H200. All axes are logarithmic. (a) Median runtime of five calls against the number of rows; shading is the interquartile range. (b) Speedup, the TabImpute median divided by the MatrixFormer median. The MatrixFormer timings were measured on an earlier checkpoint with the same architecture as the released weights, not on the released weights themselves.
  }
  \label{fig:runtime-scaling}
\end{figure}

To highlight the runtime performance of our approach, we compare the single-network MatrixFormer architecture with the entry-wise TabImpute model
\citep{feitelberg2025tabimpute} on an NVIDIA H200, fixing the width at
10 columns and increasing the row count geometrically from 16 to 1024. \Cref{fig:runtime-scaling} shows that the measured speedup increases from
$2.1\times$ at 16 rows to $107.0\times$ at 1024 rows, where median runtime
is 4.167\,s for TabImpute and 0.0389\,s for MatrixFormer.

\FloatBarrier

\FloatBarrier

\section{Synthetic priors and training objective}
\label{sec:priors-training-full}
\label{sec:prior-generation-details}

\begin{figure}[!htbp]
\centering
\includegraphics[width=1.0\linewidth]{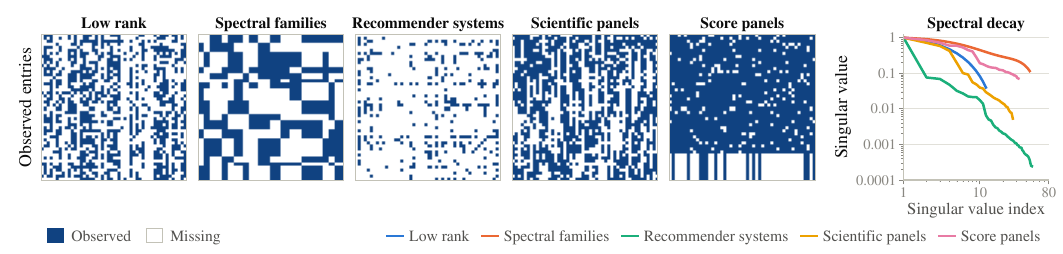}
\caption{\textbf{Synthetic training data.} Missingness masks drawn from each of MatrixFormer's five synthetic data generators, shown at a common $48 \times 48$ size (observed entries in blue, missing entries in white). Training matrices are rarely square; the recommender-system and score-panel generators train on matrices with at least 128 and 96 rows, respectively. The spectral-family example uses a neural block missing-at-random (MAR) mask, and the scientific-panel example omits its fully observed time-feature columns. Right, normalized singular values of one separate draw per generator: the raw clean signal for low rank and spectral families (slow decay), the expected rating for recommender systems, and the training target otherwise. Each line ends at its matrix's rank.}
\label{fig:synthetic-priors}
\end{figure}

\paragraph{Spectral families.}
For the three spectral families, sample a rank parameter $k$ and matrices $U\in\mathbb{R}^{R\times k}$ and $V\in\mathbb{R}^{C\times k}$ with orthonormal columns. The generator forms
\begin{equation}
T=U\operatorname{diag}(s_1,\ldots,s_k)V^\top,
\qquad A=T+\sigma_{\epsilon}E,
\qquad E_{rc}\overset{\mathrm{iid}}{\sim}\mathcal{N}(0,1).
\label{eq:spectral-prior}
\end{equation}
The noise level $\sigma_{\epsilon}$ is small (\cref{tab:prior-spectrum-parameters}), and the target is the entire clean signal $T$, rather than the noisy input or an additional top-rank truncation. Thus, ``low-rank'' refers to $T$, not necessarily to $A$. With $t_j=j-1$, the spectral coefficients are
\begin{equation}
\begin{aligned}
w_j&=(1-e^{-\omega})e^{-t_j/H},\\
g_j&=\left[w_j e^{-(t_j/h)^p}+(1-w_j)\tau(1+t_j)^{-\alpha}\right]
       e^{-\lambda(t_j/(k-1))^q},
\qquad s_j=s_{\max}\frac{g_j}{g_1}.
\end{aligned}
\label{eq:prior-spectrum}
\end{equation}
This blends a rapidly decaying head with a power-law tail and a terminal taper. The low-rank family draws $k$ uniformly from $2$--$16$ and $\alpha\sim\mathrm{Unif}(0.2,2)$; the slow- and steep-tail families draw $k$ from $2$--$\min(160,R,C)$, with $\alpha\sim\mathrm{Unif}(0,0.5)$ and $\mathrm{Unif}(1,2.5)$, respectively. All ranks are capped by the matrix dimensions. The remaining parameters are sampled independently and uniformly for each matrix from the ranges in \cref{tab:prior-spectrum-parameters}. The two broader-rank families can also draw small ranks: their names describe their tail-exponent distributions, not a guarantee of full rank. The coefficient sequence in \cref{eq:prior-spectrum} need not be monotone; because $U$ and $V$ have orthonormal columns, the singular values of $T$ are its decreasing rearrangement, followed by zeros.

\begin{table}[ht]
\centering
\caption{Shared parameters of the three spectral prior families. All intervals specify uniform distributions in the raw generator's units.}
\label{tab:prior-spectrum-parameters}
\small
\begin{tabular}{@{}lll@{}}
\toprule
Parameter & Role & Range \\
\midrule
$s_{\max}$ & Leading spectral scale & $[1,5]$ \\
$\omega$ & Initial head mixture weight & $[0,3]$ \\
$h$ & Stretched-exponential head scale & $[1,30]$ \\
$p$ & Head exponent & $[1,3]$ \\
$H$ & Head-to-tail transition scale & $[1,20]$ \\
$\tau$ & Power-law tail amplitude & $[0.2,1]$ \\
$\lambda$ & Terminal taper strength & $[0,1]$ \\
$q$ & Terminal taper exponent & $[1,4]$ \\
$\sigma_{\epsilon}$ & Entrywise observation-noise standard deviation & $[0,10^{-3}]$ \\
\bottomrule
\end{tabular}
\end{table}

\paragraph{Singular-vector geometry.}
For each axis of length $n$, choose one of the following feature constructions with probability $1/4$, independently of the other axis, and take its thin QR factor with diagonal-sign correction; this gives $U$ and $V$ in \cref{eq:spectral-prior}. Writing $G_{ij}\overset{\mathrm{iid}}{\sim}\mathcal{N}(0,1)$, the four constructions are:
\begin{enumerate}
\item \emph{Gaussian:} $F=G$, yielding a Haar-distributed orthonormal basis.
\item \emph{Coherent:} $F_{ij}=e^{a_i}G_{ij}$, with $a_i\sim\mathcal{N}(0,\eta^2)$ and $\eta\sim\mathrm{Unif}(0.25,2)$, concentrating variation in selected rows.
\item \emph{Smooth:} $F_{ij}=\cos(z_i^\top\omega_j+\phi_j)+0.03G_{ij}$, where $d$ is uniform on $\{1,\ldots,5\}$, $z_i\sim\mathcal{N}(0,I_d)$, $\omega_j\sim\mathcal{N}(0,\ell^{-2}I_d)$, $\phi_j\sim\mathrm{Unif}(0,2\pi)$, and $\log\ell\sim\mathrm{Unif}(\log0.3,\log3)$. The small Gaussian term avoids rank-deficient feature matrices.
\item \emph{Offset-bearing:} $F_{i1}=1+0.05G_{i1}$ and $F_{ij}=G_{ij}$ for $j>1$.
\end{enumerate}
This varies the spatial concentration and smoothness of the signal without prescribing a single singular-vector geometry.

\paragraph{Latent-factor.}
The fourth family instead draws $k$ uniformly from $1$--$\min(11,R,C)$ and sets
\begin{equation}
T=PQ^\top,
\qquad P\in\mathbb{R}^{R\times k},\quad Q\in\mathbb{R}^{C\times k},
\qquad A=T,
\label{eq:latent-prior}
\end{equation}
where each row of either factor independently chooses one of five equally likely distributions. Its coordinates are drawn independently from $\mathcal{N}(0,1)$, $\mathrm{Laplace}(0,1)$, Student-$t_\nu$ with $\nu$ uniform on $\{2,\ldots,10\}$, or $\tfrac12\delta_0+\tfrac12\mathcal{N}(0,9)$, or the row is drawn jointly as $D-k^{-1}\mathbf{1}$ with $D\sim\mathrm{Dirichlet}(\mathbf{1}_k)$. There is no added observation noise, feature warping, or quantization in this recipe. Unlike \cref{eq:spectral-prior}, its singular values emerge from the sampled factors rather than being specified directly, and $k$ is only an upper bound on the realized rank: for example, the centered Dirichlet draw is zero when $k=1$.

\paragraph{From matrices to episodes.}
Each family is paired with either fixed-rate MCAR masking or a mixed masking kernel, yielding eight equally weighted components. The latter samples MCAR, MAR, or MNAR mechanisms~\citep{rubin1976inference}, including structured patterns, at a nominal missing rate between 5\% and 75\%; the fixed MCAR rate is 40\%. Missingness may therefore depend on $A$, rather than always being independent of its values. To mimic rows with few observations, such as new users in a recommender system, 75\% of episodes additionally thin a random subset of rows or columns; we call these sparse-support episodes. Normalization applies the same columnwise affine map to the masked input and the clean target, with an observed-support variance estimate for sparsely observed columns. The losses below are therefore computed in those standardized units.

\paragraph{Objective.}
Each episode samples a complete synthetic matrix, a mask $M$ and a set $S\subseteq\{(r,c):M_{rc}=1\}$ of scored hidden entries: all hidden entries or, in sparse-support episodes, the hidden entries of the thinned rows or columns. Let $y_{rc}$ be the true value of a hidden entry in standardized units. The point prediction is also regressed toward a reference prediction, as in knowledge distillation~\citep{hinton2015distilling}: two earlier checkpoints of the same architecture, with frozen parameters $\psi_1,\psi_2$, receive the same masked input, and their stop-gradient mean $\bar\mu_{rc}=\operatorname{sg}\big[\tfrac12(\mu_{\psi_1,rc}+\mu_{\psi_2,rc})\big]$ is the point target. The objective is
\begin{equation}
\begin{aligned}
\mathcal{L}(\theta)&=\mathbb{E}\Bigg[\frac{1}{|S|}\sum_{(r,c)\in S}
\Big(-\log p_{\theta,rc}(y_{rc})
+2\rho\big(\mu_{\theta,rc}-\bar\mu_{rc}\big)
+0.05\,\ell_{\mathrm{CDF}}\big(z_{rc},y_{rc}\big)\Big)\Bigg],\\
\rho(a)&=
\begin{cases}
\tfrac12 a^2, & |a|\le 1,\\
|a|-\tfrac12, & |a|>1,
\end{cases}
\qquad
\ell_{\mathrm{CDF}}(z,y)=\frac{1}{O}\sum_{k=1}^{O}\Big(\sum_{j\le k}q_j(z)-\mathbf{1}\{y\le\tilde b_k\}\Big)^2,
\end{aligned}
\label{eq:training-objective}
\end{equation}
where the expectation is over episodes. The point loss $\rho$ is the Huber loss with threshold one~\citep{huber1964robust}, and $\ell_{\mathrm{CDF}}$ is a ranked probability score~\citep{epstein1969rps} with thresholds at the bin midpoints, a discrete form of the continuous ranked probability score~\citep{gneiting2007scoring}. Only $\rho$ uses the reference prediction; the likelihood and CDF terms are supervised by the true value. We use $\rho$ rather than squared error because a few extreme residuals otherwise dominated the point loss.

\paragraph{Training curriculum.}
\label{sec:training-details}
\label{sec:ratings-training}
All weight updates use synthetic matrices, and training proceeds as a curriculum. The network is first trained from random initialization on small latent-factor matrices, with 10--50 rows and 5--50 columns, using the likelihood term alone. It is then trained with \cref{eq:training-objective} on all four families, with spectral matrices of 80--320 rows and 35--100 columns; the latent-factor matrices are retained at their original sizes throughout as rehearsal~\citep{robins1995catastrophic}. Next, the matrix sizes are enlarged in steps, up to 2000 rows and 1000 columns, and a copy of the network is adapted to synthetic user--item rating matrices with observed densities between $0.004$ and $0.08$. In the final stage a single network is trained on imputation and rating episodes jointly, with a variant of \cref{eq:training-objective} whose reference predictions come from the checkpoints of the previous stage. This network is MatrixFormer: inference is one forward pass of it, and no other checkpoint is used.

\paragraph{Optimization.}
The four-family stage uses the AdamW optimizer~\citep{loshchilov2019adamw} (weight decay 0.02, gradient-norm clipping at 0.5) and runs 128k updates on batches of two matrices, with FP32 parameters and BF16 computation. The learning rate warms up over 600 updates to $10^{-5}$ and then follows cosine decay~\citep{loshchilov2017sgdr} to 5\% of that peak. The size-enlargement stage runs 5k updates at half that peak learning rate. Checkpoints are selected on fixed synthetic validation suites, with one exception: the adaptation to rating matrices is selected on an inner-validation split, 10\% of the training ratings of MovieLens 100K and Netflix small; test ratings do not enter this selection. Downstream benchmarks were inspected during development, and the UCI context choice in \cref{sec:imputation-results} is an exploratory follow-up rather than a blind validation.

\FloatBarrier

\section{Tabular imputation: MissBench and UCI}
\label{sec:tabular-appendix}
\label{sec:imputation-protocol}
\label{sec:imputation-details}

We evaluate missing-value imputation on two suites. MissBench is the OpenML-based suite used with TabImpute \citep{feitelberg2025tabimpute}: 42 datasets under eleven missingness mechanisms at the nominal 40\% setting, for 462 cases. Our local UCI-style suite, motivated by HyperImpute \citep{jarrett2022hyperimpute}, contains 12 datasets under MCAR, Col-MAR and Self-MNAR at 10\%, 20\%, 30\%, 40\% and 50\% nominal missingness with ten repeats per setting, for 1,800 cases. Neither suite has a row train/test split: all rows and columns are retained, observed entries provide context, and predictions are scored only on the masked cells. We retain every dataset and case rather than a subset selected by performance or CACTI's ten-dataset list; published CACTI scores \citep{gorla2025cacti} are not directly comparable to this no-row-holdout protocol.

\paragraph{Metric.}
RMSE denotes the columnwise standard-deviation-normalized error; lower is better. For masked entries $\mathcal{M}_j$ in column $j$,
\[
e_j = \sqrt{\frac{1}{|\mathcal{M}_j|}
\sum_{i\in\mathcal{M}_j}(x_{ij}-\hat{x}_{ij})^2},
\qquad
s_j = \operatorname{std}_{i\in\mathcal{M}_j}(x_{ij}),
\qquad
\operatorname{RMSE}_{\rm case} =
\frac{1}{|\mathcal{J}|}\sum_{j\in\mathcal{J}}\frac{e_j}{s_j},
\]
where $\mathcal{J}$ contains the columns with at least one masked entry and a non-constant masked target. The denominator uses the masked ground-truth values, which are never supplied to the imputer. Every number below first averages a method's cases within a dataset and then weights datasets equally, so that large datasets and mechanisms with many cases do not dominate.

\paragraph{Methods.}
We compare MatrixFormer with eight imputers evaluated on both suites: $k$-nearest-neighbor imputation (K-NN) \citep{troyanskaya2001missing}; SoftImpute \citep{mazumder2010spectral}, a low-rank completion by iterative soft-thresholded SVD; the chained-equation methods MICE \citep{vanbuuren2011mice} and ICE, which regress each column on the others in turn; MissForest \citep{stekhoven2012missforest}, which does the same with random forests; the deep generative models GAIN \citep{yoon2018gain} and MIWAE \citep{mattei2019miwae}; and HyperImpute \citep{jarrett2022hyperimpute}, which selects a model for each column within an iterative scheme. The baselines reuse saved predictions without retraining. MatrixFormer uses whole-matrix inference on MissBench. On UCI it uses sequential windows of at most 2,000 rows and 1,000 columns, the largest matrices seen in pre-training, with observed-column normalization computed once from the entire incomplete matrix and overlapping tail-window predictions averaged; only rows are split, since these datasets have at most 90 columns. This window limit was adopted after inspecting whole-matrix results, so it is an exploratory protocol choice.

\subsection{MissBench}
\label{sec:missbench-appendix}

\paragraph{Datasets and mechanisms.}
MissBench consists of 42 OpenML datasets (\cref{tab:missbench-datasets}). They are small: between 50 and 235 rows, with a median of 98, and between 5 and 55 columns, with a median of 10. Each dataset is masked by eleven missingness mechanisms at the nominal 40\% setting, giving 462 cases: MCAR, Col-MAR, Self-MNAR, neural MAR, block-neural MAR, and six structured MNAR mechanisms (polarization, soft polarization, latent factor, cluster level, two-phase subset and censoring). The structured mechanisms need not realize exactly 40\% missing entries. The generators are those of \citet{feitelberg2025tabimpute}; Col-MAR and Self-MNAR, which UCI also uses, are defined in \cref{sec:uci-appendix}.

\begin{table}[!htbp]
\centering
\footnotesize
\setlength{\tabcolsep}{3.5pt}
\caption{The 42 MissBench datasets, all from OpenML, with the number of rows and columns of each matrix.}
\label{tab:missbench-datasets}
\resizebox{\ifdim\width>\linewidth\linewidth\else\width\fi}{!}{%
\begin{tabular}{@{}lrr@{\qquad}lrr@{\qquad}lrr@{}}
\toprule
Dataset & Rows & Cols & Dataset & Rows & Cols & Dataset & Rows & Cols \\
\midrule
alcohol-qcm-sensor & 125 & 15 & edm & 154 & 18 & rabe\_265 & 51 & 7 \\
analcatdata\_election2000 & 67 & 15 & EgyptianSkulls & 150 & 5 & Rainfall-in-Kerala-1901-2017 & 117 & 18 \\
analcatdata\_gviolence & 74 & 9 & FacultySalaries & 50 & 5 & sleuth\_case1201 & 50 & 7 \\
analcatdata\_olympic2000 & 66 & 12 & Glass-Classification & 214 & 10 & sleuth\_ex1605 & 62 & 6 \\
appendicitis\_test & 106 & 8 & humans\_numeric & 75 & 15 & slump & 103 & 10 \\
appendicitis\_test\_edsa & 106 & 8 & ICU & 200 & 20 & SMSA & 59 & 16 \\
autoPrice & 159 & 16 & machine\_cpu & 209 & 7 & SolarPower & 204 & 5 \\
baskball & 96 & 5 & MercuryinBass & 53 & 10 & Student-Scores & 56 & 13 \\
benzo32 & 195 & 33 & metafeatures & 75 & 32 & Swiss-banknote-conterfeit-detection & 200 & 7 \\
breast-cancer-coimbra & 116 & 10 & pollution & 60 & 16 & treepipit & 86 & 10 \\
chatfield\_4 & 235 & 13 & pwLinear & 200 & 11 & visualizing\_hamster & 73 & 6 \\
chscase\_vine1 & 52 & 10 & pyrim & 74 & 28 & WineDataset & 178 & 14 \\
dataset\_analcatdata\_creditscore & 100 & 7 & rabe\_131 & 50 & 6 & wisconsin & 194 & 33 \\
divorce\_prediction & 170 & 55 & rabe\_148 & 66 & 6 & witmer\_census\_1980 & 50 & 5 \\
\bottomrule
\end{tabular}}
\end{table}

\paragraph{Results by mechanism.}
\Cref{tab:missbench-mechanisms} separates the overall score of \cref{fig:empirical-overview}(a) into the eleven mechanisms. MatrixFormer has the lowest error in nine. K-NN is lower on Col-MAR ($1.075$ against $1.105$) and on block-neural MAR ($1.039$ against $1.061$). The mechanisms differ greatly in difficulty: under censoring and polarization every method has an error above $3$, and these two mechanisms account for much of each method's overall score.

\begin{table}[!htbp]
\centering
\footnotesize
\setlength{\tabcolsep}{3.5pt}
\caption{MissBench RMSE by missingness mechanism at the nominal 40\% setting; lower is better. Each entry averages the 42 datasets; Overall first averages the eleven mechanisms within each dataset. Neural MAR and Block MAR are the neural and block-neural MAR mechanisms; the six columns from Polar.\ to Censoring are the structured MNAR mechanisms. The lowest value in each column is in bold.}
\label{tab:missbench-mechanisms}
\resizebox{\ifdim\width>\linewidth\linewidth\else\width\fi}{!}{%
\begin{tabular}{@{}lrrrrrrrrrrrr@{}}
\toprule
Method & MCAR & Col-MAR & Self-MNAR & Neural MAR & Block MAR & Polar. & Soft polar. & Latent & Cluster & Two-phase & Censoring & Overall \\
\midrule
MatrixFormer & \textbf{0.861} & 1.105 & \textbf{1.950} & \textbf{0.871} & 1.061 & \textbf{3.479} & \textbf{2.267} & \textbf{0.909} & \textbf{0.881} & \textbf{0.908} & \textbf{3.437} & \textbf{1.612} \\
MissForest & 1.018 & 1.197 & 2.580 & 1.070 & 1.430 & 4.433 & 2.498 & 1.052 & 0.947 & 0.937 & 4.465 & 1.966 \\
K-NN & 0.943 & \textbf{1.075} & 2.539 & 1.004 & \textbf{1.039} & 3.660 & 3.157 & 0.958 & 0.988 & 0.915 & 5.961 & 2.022 \\
HyperImpute & 0.969 & 1.117 & 2.636 & 1.114 & 1.325 & 5.492 & 2.713 & 1.082 & 0.917 & 0.931 & 4.444 & 2.067 \\
MICE & 1.002 & 1.298 & 2.543 & 1.061 & 1.751 & 6.231 & 2.494 & 1.062 & 0.977 & 0.931 & 5.051 & 2.218 \\
ICE & 1.215 & 1.135 & 2.988 & 1.171 & 1.299 & 6.318 & 2.829 & 1.181 & 1.068 & 0.947 & 4.762 & 2.265 \\
SoftImpute & 1.007 & 1.417 & 3.372 & 1.270 & 1.243 & 4.750 & 2.804 & 1.056 & 1.079 & 1.198 & 5.978 & 2.289 \\
MIWAE & 1.235 & 1.701 & 3.608 & 1.246 & 1.384 & 4.044 & 2.753 & 1.301 & 1.224 & 1.322 & 6.284 & 2.373 \\
GAIN & 1.246 & 2.207 & 2.965 & 1.667 & 2.484 & 8.751 & 2.287 & 1.833 & 1.348 & 1.711 & 7.222 & 3.065 \\
\bottomrule
\end{tabular}}
\end{table}

\paragraph{Results by dataset.}
\Cref{tab:missbench-per-dataset} gives the error on each dataset, averaged over the eleven mechanisms. MatrixFormer has the lowest error on 31 of the 42 datasets and is among the two lowest on 36. It is not uniformly best: it ranks eighth of nine on ICU and last on treepipit, and fourth on three further datasets.

\begin{table}[p]
\centering
\footnotesize
\setlength{\tabcolsep}{3.5pt}
\caption{MissBench RMSE by dataset, averaged over the eleven missingness mechanisms; lower is better. The lowest value in each row is in bold.}
\label{tab:missbench-per-dataset}
\resizebox{\ifdim\width>\linewidth\linewidth\else\width\fi}{!}{%
\begin{tabular}{@{}lrrrrrrrrr@{}}
\toprule
Dataset & MatrixFormer & MissForest & K-NN & HyperImpute & MICE & ICE & SoftImpute & MIWAE & GAIN \\
\midrule
alcohol-qcm-sensor & \textbf{0.782} & 1.205 & 1.805 & 0.945 & 1.083 & 0.946 & 1.973 & 2.442 & 2.422 \\
analcatdata\_election2000 & \textbf{2.811} & 3.985 & 9.602 & 8.657 & 11.42 & 8.878 & 12.77 & 7.576 & 12.78 \\
analcatdata\_gviolence & \textbf{1.498} & 1.877 & 2.048 & 1.933 & 1.885 & 2.078 & 2.038 & 1.958 & 2.888 \\
analcatdata\_olympic2000 & \textbf{1.087} & 1.303 & 1.602 & 1.184 & 1.280 & 1.274 & 1.864 & 2.293 & 2.054 \\
appendicitis\_test & \textbf{0.938} & 1.183 & 1.275 & 0.956 & 1.055 & 1.033 & 1.316 & 1.465 & 1.795 \\
appendicitis\_test\_edsa & \textbf{0.908} & 1.397 & 1.493 & 0.959 & 1.295 & 0.992 & 1.675 & 1.816 & 1.656 \\
autoPrice & \textbf{1.231} & 1.475 & 1.623 & 1.561 & 1.370 & 1.537 & 1.574 & 1.831 & 2.335 \\
baskball & \textbf{1.572} & 1.763 & 1.662 & 1.701 & 1.642 & 1.694 & 1.871 & 1.717 & 1.783 \\
benzo32 & \textbf{1.141} & 1.366 & 1.349 & 1.388 & 1.368 & 1.954 & 1.277 & 1.439 & 2.424 \\
breast-cancer-coimbra & \textbf{1.652} & 2.081 & 2.118 & 2.120 & 2.172 & 2.172 & 2.325 & 2.613 & 2.341 \\
chatfield\_4 & \textbf{0.890} & 1.322 & 1.766 & 1.052 & 1.368 & 1.039 & 1.617 & 2.397 & 1.570 \\
chscase\_vine1 & \textbf{1.465} & 1.545 & 1.563 & 1.573 & 1.468 & 1.661 & 1.644 & 1.915 & 2.312 \\
dataset\_analcatdata\_creditscore & 1.968 & 1.855 & 2.060 & 2.003 & 1.988 & 2.309 & \textbf{1.818} & 2.191 & 1.967 \\
divorce\_prediction & \textbf{1.009} & 1.024 & 1.076 & 1.048 & 1.044 & 1.126 & 1.127 & 1.849 & 1.511 \\
edm & 1.174 & 1.181 & 1.182 & \textbf{1.053} & 1.085 & 1.097 & 1.330 & 1.789 & 1.942 \\
EgyptianSkulls & \textbf{1.461} & 1.634 & 1.628 & 1.627 & 1.624 & 1.617 & 1.494 & 1.681 & 1.666 \\
FacultySalaries & \textbf{0.907} & 1.238 & 1.625 & 1.365 & 1.331 & 1.329 & 1.648 & 2.178 & 1.728 \\
Glass-Classification & 1.244 & 1.360 & 1.423 & 1.299 & 1.320 & \textbf{1.228} & 1.527 & 1.903 & 1.969 \\
humans\_numeric & \textbf{2.542} & 3.271 & 3.252 & 3.370 & 3.666 & 4.130 & 2.974 & 3.593 & 3.330 \\
ICU & 2.176 & 1.778 & 1.796 & 1.998 & 1.749 & 2.275 & \textbf{1.708} & 1.933 & 1.779 \\
machine\_cpu & \textbf{1.871} & 2.417 & 2.504 & 2.359 & 2.580 & 2.933 & 3.297 & 3.886 & 3.014 \\
MercuryinBass & \textbf{1.367} & 1.591 & 1.810 & 1.573 & 1.486 & 1.639 & 2.166 & 2.230 & 1.716 \\
metafeatures & 8.606 & 13.62 & \textbf{7.794} & 14.54 & 18.26 & 19.04 & 10.82 & 11.27 & 30.39 \\
pollution & \textbf{1.445} & 1.690 & 1.792 & 1.805 & 1.686 & 1.973 & 2.185 & 1.956 & 3.069 \\
pwLinear & 1.351 & 1.735 & 1.422 & 1.717 & \textbf{1.301} & 1.630 & 1.373 & 1.596 & 1.627 \\
pyrim & 1.360 & 1.320 & 1.349 & 1.367 & \textbf{1.234} & 1.482 & 1.424 & 1.660 & 1.634 \\
rabe\_131 & \textbf{1.545} & 1.694 & 1.717 & 1.601 & 1.611 & 1.598 & 1.750 & 2.000 & 2.215 \\
rabe\_148 & \textbf{1.387} & 1.598 & 1.573 & 1.686 & 1.577 & 1.709 & 1.889 & 1.949 & 2.053 \\
rabe\_265 & \textbf{1.546} & 1.881 & 1.832 & 1.981 & 1.821 & 2.054 & 2.029 & 1.795 & 2.799 \\
Rainfall-in-Kerala-1901-2017 & \textbf{1.459} & 1.787 & 1.772 & 1.594 & 1.712 & 1.580 & 1.706 & 1.939 & 2.271 \\
sleuth\_case1201 & 1.470 & 1.611 & 1.490 & \textbf{1.453} & 1.470 & 1.480 & 1.975 & 1.822 & 2.283 \\
sleuth\_ex1605 & \textbf{1.432} & 1.710 & 1.656 & 1.677 & 1.609 & 1.668 & 1.633 & 1.837 & 2.138 \\
slump & \textbf{1.765} & 1.968 & 1.990 & 1.833 & 1.929 & 1.846 & 2.060 & 2.276 & 2.656 \\
SMSA & \textbf{1.637} & 1.841 & 1.833 & 2.054 & 1.834 & 2.024 & 2.242 & 1.953 & 2.830 \\
SolarPower & 1.546 & 1.751 & 1.742 & 1.675 & 1.688 & \textbf{1.528} & 2.009 & 1.913 & 1.813 \\
Student-Scores & \textbf{0.764} & 1.245 & 1.436 & 1.102 & 1.307 & 1.174 & 1.959 & 2.098 & 2.263 \\
Swiss-banknote-conterfeit-detection & 1.249 & 1.339 & 1.322 & 1.276 & \textbf{1.221} & 1.279 & 1.581 & 1.796 & 1.795 \\
treepipit & 2.520 & 2.223 & \textbf{2.102} & 2.253 & 2.191 & 2.478 & 2.223 & 2.511 & 2.203 \\
visualizing\_hamster & \textbf{1.373} & 1.552 & 1.515 & 1.559 & 1.427 & 1.523 & 1.927 & 1.742 & 1.696 \\
WineDataset & \textbf{1.274} & 1.364 & 1.362 & 1.417 & 1.280 & 1.491 & 1.602 & 1.765 & 2.052 \\
wisconsin & \textbf{0.901} & 1.174 & 1.411 & 0.955 & 1.221 & 1.018 & 1.205 & 1.441 & 1.928 \\
witmer\_census\_1980 & \textbf{1.367} & 1.623 & 1.538 & 1.555 & 1.503 & 1.604 & 1.496 & 1.647 & 2.044 \\
\bottomrule
\end{tabular}}
\end{table}

\subsection{UCI}
\label{sec:uci-appendix}

\paragraph{Datasets and mechanisms.}
The UCI suite, motivated by HyperImpute \citep{jarrett2022hyperimpute}, has twelve datasets (\cref{tab:uci-datasets}), from 150 to 20,640 rows and from 5 to 90 columns. Each is masked under MCAR, Col-MAR and Self-MNAR at 10\%, 20\%, 30\%, 40\% and 50\% nominal missingness with ten repeats, giving 1,800 cases. The Col-MAR and Self-MNAR generators are the same as in MissBench.
Col-MAR keeps a randomly selected subset of predictor columns fully observed.
A random linear combination $z_i$ of those observed values determines the
missingness probability for each entry in the remaining columns:
$\sigma(z_i-Q_{1-p}(z))$, with independent Bernoulli draws across entries.
Self-MNAR instead selects a random nonempty subset of target columns and masks
each target entry with probability
$\sigma(a_j x_{ij}-Q_{1-p}(a_j X_{\cdot j}))$, where
$a_j\in\{-2,-1,1,2\}$ is drawn independently for each target column.
Thus its masking probability depends on the entry's own value; unselected
columns remain observed. Here $\sigma$ is the logistic function and $Q_{1-p}$
is an empirical quantile. The parameter $p$ is a nominal setting, not a
constraint on the realized whole-matrix missing fraction.

\begin{table}[!htbp]
\centering
\small
\setlength{\tabcolsep}{3.5pt}
\caption{The twelve UCI datasets, with the number of rows and columns of each matrix.}
\label{tab:uci-datasets}
\resizebox{\ifdim\width>\linewidth\linewidth\else\width\fi}{!}{%
\begin{tabular}{@{}lrr@{\qquad}lrr@{}}
\toprule
Dataset & Rows & Cols & Dataset & Rows & Cols \\
\midrule
Airfoil Self-Noise & 1,503 & 5 & Landsat Satellite & 6,435 & 36 \\
California Housing & 20,640 & 9 & Letter Recognition & 20,000 & 16 \\
Concrete Compression & 1,030 & 8 & Libras Movement & 360 & 90 \\
Diabetes & 442 & 11 & Spam Base & 4,601 & 57 \\
Ionosphere & 351 & 34 & Wine Quality (red) & 1,599 & 11 \\
Iris & 150 & 5 & Wine Quality (white) & 4,898 & 11 \\
\bottomrule
\end{tabular}}
\end{table}

\paragraph{Coverage.}
MatrixFormer, MissForest, K-NN, MICE, ICE and SoftImpute have predictions for all 1,800 cases. GAIN, HyperImpute and MIWAE have 1,556: they lack 244 of the 600 MCAR cases, on seven datasets. Their entries marked $\dagger$ average the cases they have, so they are not computed on the same cases as the other methods.

\paragraph{Results by mechanism and rate.}
MatrixFormer has the lowest error under each of the three mechanisms and at each of the five missingness rates (\cref{tab:uci-mechanisms-rates}). Its margin is small under MCAR ($0.689$ against $0.694$ for SoftImpute) and Col-MAR ($0.826$ against $0.841$ for ICE) and large under Self-MNAR ($1.536$ against $1.952$).

\begin{table}[!htbp]
\centering
\small
\setlength{\tabcolsep}{3.5pt}
\caption{UCI RMSE by missingness mechanism (averaged over the five rates) and by nominal missingness rate (averaged over the three mechanisms); lower is better. Each entry averages ten repeats within a dataset and then the twelve datasets. $\dagger$: the method lacks some MCAR cases and is averaged over its valid cases. The lowest value in each column is in bold.}
\label{tab:uci-mechanisms-rates}
\resizebox{\ifdim\width>\linewidth\linewidth\else\width\fi}{!}{%
\begin{tabular}{@{}lrrr@{\quad}rrrrr@{\quad}r@{}}
\toprule
 & \multicolumn{3}{c}{Mechanism} & \multicolumn{5}{c}{Missingness rate} & \\
\cmidrule(lr){2-4}\cmidrule(lr){5-9}
Method & MCAR & Col-MAR & Self-MNAR & 10\% & 20\% & 30\% & 40\% & 50\% & Overall \\
\midrule
MatrixFormer & \textbf{0.689} & \textbf{0.826} & \textbf{1.536} & \textbf{1.171} & \textbf{0.985} & \textbf{0.996} & \textbf{0.964} & \textbf{0.969} & \textbf{1.017} \\
MissForest & 0.702 & 0.864 & 2.025 & 1.418 & 1.150 & 1.171 & 1.105 & 1.142 & 1.197 \\
K-NN & 0.748 & 0.853 & 2.007 & 1.334 & 1.146 & 1.198 & 1.124 & 1.211 & 1.202 \\
HyperImpute & 0.710$^\dagger$ & 0.957 & 1.955 & 1.516$^\dagger$ & 1.302$^\dagger$ & 1.313$^\dagger$ & 1.084 & 1.372$^\dagger$ & 1.301$^\dagger$ \\
MICE & 1.131 & 1.248 & 2.494 & 2.034 & 1.556 & 1.635 & 1.402 & 1.496 & 1.625 \\
ICE & 0.722 & 0.841 & 1.952 & 1.325 & 1.097 & 1.138 & 1.100 & 1.198 & 1.172 \\
SoftImpute & 0.694 & 0.917 & 2.187 & 1.495 & 1.230 & 1.236 & 1.178 & 1.193 & 1.266 \\
MIWAE & 3.445$^\dagger$ & 1.229 & 2.560 & 2.579$^\dagger$ & 2.121$^\dagger$ & 2.016$^\dagger$ & 2.247 & 1.813$^\dagger$ & 2.168$^\dagger$ \\
GAIN & 0.941$^\dagger$ & 1.715 & 2.661 & 2.703$^\dagger$ & 1.841$^\dagger$ & 2.032$^\dagger$ & 1.601 & 1.663$^\dagger$ & 1.950$^\dagger$ \\
\bottomrule
\end{tabular}}
\end{table}

\paragraph{Results by dataset.}
MatrixFormer has the lowest error on seven of the twelve datasets and is among the two lowest on ten (\cref{tab:uci-per-dataset}). It ranks fourth on Iris and fifth on Libras Movement, where the chained-equation methods have less than half its error.

\begin{table}[!htbp]
\centering
\footnotesize
\setlength{\tabcolsep}{3.5pt}
\caption{UCI RMSE by dataset, averaged over the three mechanisms, five rates and ten repeats; lower is better. $\dagger$: the method lacks some MCAR cases on this dataset. The lowest value in each row is in bold.}
\label{tab:uci-per-dataset}
\resizebox{\ifdim\width>\linewidth\linewidth\else\width\fi}{!}{%
\begin{tabular}{@{}lrrrrrrrrr@{}}
\toprule
Dataset & MatrixFormer & MissForest & K-NN & HyperImpute & MICE & ICE & SoftImpute & MIWAE & GAIN \\
\midrule
Airfoil Self-Noise & \textbf{1.884} & 1.998 & 2.051 & 2.075$^\dagger$ & 2.856 & 2.049 & 2.311 & 2.999$^\dagger$ & 3.404$^\dagger$ \\
California Housing & \textbf{1.137} & 1.397 & 1.464 & 1.766 & 2.683 & 1.496 & 1.355 & 3.066 & 3.370 \\
Concrete Compression & \textbf{1.792} & 2.227 & 2.262 & 2.200 & 2.598 & 1.876 & 2.211 & 3.587 & 2.531 \\
Diabetes & \textbf{0.834} & 0.927 & 0.977 & 0.908$^\dagger$ & 1.082 & 0.885 & 0.889 & 1.864$^\dagger$ & 1.168$^\dagger$ \\
Ionosphere & 0.802 & 0.836 & \textbf{0.792} & 0.846 & 1.110 & 1.005 & 0.882 & 1.265 & 0.983 \\
Iris & 0.584 & \textbf{0.554} & 0.630 & 0.570$^\dagger$ & 0.778 & 0.555 & 0.891 & 1.940$^\dagger$ & 0.831$^\dagger$ \\
Landsat Satellite & \textbf{0.625} & 1.357 & 1.018 & 1.513$^\dagger$ & 0.694 & 0.785 & 1.515 & 3.265$^\dagger$ & 3.054$^\dagger$ \\
Letter Recognition & 1.093 & 1.200 & \textbf{0.985} & 1.115 & 1.126 & 1.136 & 1.279 & 1.509 & 1.556 \\
Libras Movement & 0.184 & 0.324 & 0.424 & 0.093 & \textbf{0.084} & 0.092 & 0.138 & 1.694 & 0.474 \\
Spam Base & 1.238 & 1.264 & 1.559 & 2.025$^\dagger$ & 3.145 & 2.043 & 1.337 & \textbf{1.235}$^\dagger$ & 2.707$^\dagger$ \\
Wine Quality (red) & \textbf{1.023} & 1.150 & 1.137 & 1.309$^\dagger$ & 1.448 & 1.073 & 1.211 & 2.075$^\dagger$ & 1.726$^\dagger$ \\
Wine Quality (white) & \textbf{1.008} & 1.131 & 1.133 & 1.196$^\dagger$ & 1.891 & 1.064 & 1.174 & 1.521$^\dagger$ & 1.599$^\dagger$ \\
\bottomrule
\end{tabular}}
\end{table}

\FloatBarrier

\FloatBarrier

\section{Causal panel data: counterfactual completion}
\label{sec:causal-protocol}
\label{sec:causal-details}

Estimating a treatment effect from panel data amounts to predicting the untreated outcomes of the treated units in the post-treatment periods from the remaining entries \citep{athey2021matrix}. We evaluate MatrixFormer on 3,000 saved placebo draws in 12 designs, where the untreated values are known and the true effect is zero. The benchmarks are the CPS placebo study of \citet{arkhangelsky2021synthetic}\footnote{state-by-year panels simulated from Current Population Survey log wages, hours and unemployment rates, with treatment assigned to mimic state minimum-wage, abortion-ban and open-carry laws, all nine combinations}, the Kansas semi-synthetic factor design of
\citet{benmichael2021augmented} on one treated state, and the
California-smoking donor pool \citep{abadie2010synthetic} with simultaneous
or staggered pseudo-treatment following \citet{athey2021matrix}.
All three benchmarks start from real panels. CPS and Kansas simulate new panels from models fitted to the real data; California smoking uses the real outcomes directly and only the treatment is artificial. In every draw the hidden cells are untreated outcomes, so the true treatment effect is zero.

\paragraph{CPS placebo panels.}
The source is the state-by-year Current Population Survey panel shipped with the \texttt{synthdid} R package (version 0.0.9, \texttt{sdid-paper} branch):
50 states and 40 years, with outcome log wage, hours worked, or unemployment rate. Following \citet{arkhangelsky2021synthetic}, the outcome matrix $Y$ is standardized by its overall mean and standard deviation. Its rank-4 SVD approximation $L$ is split into an additive part $F$ (the unit and time means of $L$) and an interactive part $M=L-F$, and $E=Y-L$ is the residual. Simulated panels are $Y^{*}=F+M+\varepsilon$, where each state's row $\varepsilon_i\sim\mathcal{N}(0,\Sigma)$ and $\Sigma$ is an AR(2) correlation matrix over years, with coefficients fitted to $E$ by least squares and scale matched to the covariance of $E$. For treatment, each state's indicator for the chosen policy (minimum wage, abortion ban, or open carry) in the final year is regressed on the four unit factors by logistic regression, giving probabilities $\pi_i$. Each draw assigns states independently with probability $\pi_i$ and keeps at most 10 of them at random (10 random states if none are drawn). The last 10 years are post-treatment ($T_0=30$), and the hidden block is the treated states in those years. We generate 200 draws for each of the nine outcome--assignment combinations. Outcomes are on the standardized scale.

\paragraph{Kansas semi-synthetic panels.}
The source is the \texttt{augsynth::kansas} (R) panel of quarterly log state GDP per capita: 50 states and 105 quarters (1990 Q1--2016 Q1), with 89 quarters before the 2012 Kansas tax cut and 16 after it (from 2012 Q2). As in the main factor-model simulation of \citet{benmichael2021augmented}, we fit the two-way interactive fixed-effects model $y_{it}=\alpha_i+\xi_t+\phi_i^\top\mu_t+\varepsilon_{it}$ with three factors.
Draws keep the time effects $\hat\xi_t$ and factors $\hat\mu_t$ fixed and resample $\alpha_i\sim\mathcal{N}(\bar{\hat\alpha},\mathrm{sd}(\hat\alpha))$, $\phi_i\sim\mathcal{N}(0,\mathrm{Cov}(\hat\phi))$ and $\varepsilon_{it}\sim\mathcal{N}(0,\hat\sigma^2)$ independently. One treated state is sampled with probability proportional to $\operatorname{logit}^{-1}\bigl(\theta\,(z(\alpha_i)+\textstyle\sum_j z(\phi_{ij}))\bigr)$ with $\theta=0.5$, where $z$ standardizes across states. Its 16 post-treatment quarters are hidden. There are 200 draws.
\paragraph{California smoking panels.}
These are real data with no simulation: annual cigarette sales in packs per capita for 1970--2000 ($T=31$) in the 38 states of the Proposition 99 donor pool \citep{abadie2010synthetic}. California is excluded, so no state is actually treated. Following \citet{athey2021matrix}, $T_0=\operatorname{round}(\rho T)$ with $\rho\in\{0.35,0.5,0.65,0.8,0.9\}$, i.e.\ $T_0\in\{11,16,20,25,28\}$, and 100 draws per value. In the simultaneous design, 8 random states are hidden from year $T_0+1$ onward. In the staggered design, 35 random states are each hidden from an adoption year drawn uniformly after the first $T_0$ years.

\paragraph{Metrics.}
For draw $d$ with hidden block $\mathcal{H}_d$, the treated units in their post-treatment periods, true untreated outcomes
$y_{it}$ and predictions $\hat y_{it}$,
\[
\operatorname{RMSE}_d = \sqrt{\frac{1}{|\mathcal{H}_d|}\sum_{(i,t)\in\mathcal{H}_d}(y_{it}-\hat y_{it})^2},
\qquad
|\hat\tau_d-\tau_d| = \Bigl|\frac{1}{|\mathcal{H}_d|}\sum_{(i,t)\in\mathcal{H}_d}(y_{it}-\hat y_{it})\Bigr|.
\]
The second is the absolute error of the implied average treatment effect on the treated: the hidden entries are untreated, so $\tau_d=0$ and the error equals the absolute mean bias of the completion. Both metrics are in the outcome's units. We average each over the draws of a design, and a benchmark's value in \cref{fig:empirical-overview}(c) and (d) gives its designs equal weight. With per-design standard errors $\mathrm{se}_g$ over draws and $G$ designs, the interval is $\pm 1.96\,\sqrt{\sum_g \mathrm{se}_g^2}/G$. The intervals treat the designs as fixed and the draws as independent; draws share their underlying source data, so the intervals may be somewhat optimistic.

\paragraph{Methods.}
The comparison includes the difference-in-differences (DiD), synthetic control (SC) and synthetic difference-in-differences (SDiD) estimators of \citet{arkhangelsky2021synthetic}, and the SoftImpute, K-NN and HyperImpute \citep{jarrett2022hyperimpute} imputers. It also includes two cross-sectional effect estimators that are not designed for panel data, the T-learner \citep{kunzel2019metalearners} with random forests and the causal foundation model CausalPFN \citep{balazadeh2025causalpfn}: each is fit separately in every post-treatment period with a unit's pre-treatment outcomes as its covariates, without tuning; its RMSE uses its untreated-outcome model fit on the control units only, and its effect error uses its own effect estimate. All methods use the same saved draws and hidden blocks.

MatrixFormer uses the same fixed checkpoint as in the other domains, with no training or checkpoint selection for this evaluation. Each panel has units as rows and time periods as columns. Column means and standard deviations use only observed entries. MatrixFormer completes the entire standardized panel in one forward pass, using the predictive mean; we then undo the standardization and restore every observed entry.

\paragraph{Results by design.}
At the level of a benchmark, MatrixFormer has the lowest RMSE and the lowest effect error on all three (\cref{fig:empirical-overview}c and d), by a small margin on CPS (RMSE $0.532$ against $0.539$ for SDiD, effect error $0.093$ against $0.096$). \Cref{tab:causal-design-rmse,tab:causal-design-effect} separate the twelve designs. MatrixFormer has the lowest value in four of the twelve for each metric; it is first or second in every design for RMSE and within the first three for effect error. SDiD has the lowest RMSE in six designs.

The comparison with SDiD depends on the design. In paired comparisons over draws, with $95\%$ normal intervals for the mean difference, MatrixFormer has lower RMSE in five designs (the three CPS unemployment designs, Kansas and staggered California), SDiD has lower RMSE in the three CPS log-wage designs, and the remaining four do not separate. For effect error the corresponding counts are four, two and six. The largest difference is under staggered adoption in California, where MatrixFormer's RMSE is $13.19$ against $16.13$ and its effect error $3.08$ against $5.71$; under simultaneous adoption the two methods do not separate. MatrixFormer's lead on the California benchmark therefore comes from the staggered design, and its RMSE lead on CPS from the unemployment designs.

\begin{table}[!htbp]
\centering
\footnotesize
\setlength{\tabcolsep}{3.5pt}
\caption{Treated-block RMSE by placebo design, in the outcome's units; lower is better. Each entry is a mean over the draws of the design (200 per CPS and Kansas design, 500 per California design). The lowest value in each row is in bold.}
\label{tab:causal-design-rmse}
\resizebox{\linewidth}{!}{%
\begin{tabular}{@{}lrrrrrrrrr@{}}
\toprule
Design & MatrixFormer & SDiD & SC & DiD & SoftImpute & K-NN & HyperImpute & T-learner & CausalPFN \\
\midrule
\multicolumn{10}{@{}l}{\emph{CPS}} \\
\quad hours, abortion ban & \textbf{0.720} & 0.725 & 0.838 & 0.846 & 0.767 & 0.783 & 0.776 & 0.768 & 0.755 \\
\quad hours, minimum wage & 0.754 & \textbf{0.753} & 0.834 & 0.839 & 0.782 & 0.822 & 0.825 & 0.805 & 0.790 \\
\quad hours, open carry & 0.724 & \textbf{0.720} & 0.811 & 0.817 & 0.770 & 0.777 & 0.777 & 0.764 & 0.763 \\
\quad log wage, abortion ban & 0.126 & \textbf{0.121} & 0.167 & 0.169 & 0.145 & 0.152 & 0.145 & 0.141 & 0.134 \\
\quad log wage, minimum wage & 0.126 & \textbf{0.119} & 0.152 & 0.156 & 0.137 & 0.161 & 0.151 & 0.147 & 0.135 \\
\quad log wage, open carry & 0.125 & \textbf{0.121} & 0.155 & 0.157 & 0.131 & 0.138 & 0.136 & 0.133 & 0.133 \\
\quad unemployment, abortion ban & 0.729 & 0.754 & 0.796 & 0.807 & \textbf{0.719} & 0.791 & 0.781 & 0.773 & 0.764 \\
\quad unemployment, minimum wage & 0.742 & 0.771 & 0.828 & 0.861 & \textbf{0.734} & 0.811 & 0.807 & 0.804 & 0.763 \\
\quad unemployment, open carry & \textbf{0.740} & 0.765 & 0.796 & 0.803 & 0.744 & 0.804 & 0.790 & 0.778 & 0.788 \\
\midrule
\multicolumn{10}{@{}l}{\emph{California smoking}} \\
\quad simultaneous & 10.22 & \textbf{10.00} & 15.84 & 16.64 & 14.40 & 15.86 & 13.19 & 12.96 & 11.54 \\
\quad staggered & \textbf{13.19} & 16.13 & 19.74 & 20.40 & 18.55 & 21.66 & 16.51 & 19.14 & 17.94 \\
\midrule
\multicolumn{10}{@{}l}{\emph{Kansas}} \\
\quad factor model & \textbf{0.027} & 0.030 & 0.040 & 0.086 & 0.059 & 0.071 & 0.030 & 0.052 & 0.033 \\
\bottomrule
\end{tabular}}
\end{table}

\begin{table}[!htbp]
\centering
\footnotesize
\setlength{\tabcolsep}{3.5pt}
\caption{Absolute treatment-effect error $|\hat\tau-\tau|$ by placebo design, in the outcome's units; lower is better. Layout as in \cref{tab:causal-design-rmse}.}
\label{tab:causal-design-effect}
\resizebox{\linewidth}{!}{%
\begin{tabular}{@{}lrrrrrrrrr@{}}
\toprule
Design & MatrixFormer & SDiD & SC & DiD & SoftImpute & K-NN & HyperImpute & T-learner & CausalPFN \\
\midrule
\multicolumn{10}{@{}l}{\emph{CPS}} \\
\quad hours, abortion ban & \textbf{0.114} & 0.123 & 0.165 & 0.188 & 0.131 & 0.137 & 0.130 & 0.129 & 0.126 \\
\quad hours, minimum wage & \textbf{0.143} & 0.154 & 0.148 & 0.157 & 0.165 & 0.202 & 0.239 & 0.172 & 0.210 \\
\quad hours, open carry & 0.121 & \textbf{0.120} & 0.146 & 0.165 & 0.142 & 0.144 & 0.156 & 0.129 & 0.140 \\
\quad log wage, abortion ban & 0.028 & \textbf{0.020} & 0.029 & 0.036 & 0.068 & 0.058 & 0.043 & 0.044 & 0.032 \\
\quad log wage, minimum wage & 0.035 & \textbf{0.021} & 0.032 & 0.043 & 0.063 & 0.083 & 0.068 & 0.068 & 0.048 \\
\quad log wage, open carry & 0.021 & 0.022 & 0.030 & 0.034 & 0.026 & 0.026 & 0.024 & 0.025 & \textbf{0.021} \\
\quad unemployment, abortion ban & 0.099 & 0.112 & 0.162 & 0.191 & \textbf{0.091} & 0.139 & 0.129 & 0.127 & 0.106 \\
\quad unemployment, minimum wage & 0.173 & 0.193 & 0.304 & 0.361 & \textbf{0.148} & 0.243 & 0.257 & 0.252 & 0.166 \\
\quad unemployment, open carry & 0.107 & 0.102 & 0.130 & 0.149 & \textbf{0.097} & 0.124 & 0.132 & 0.109 & 0.130 \\
\midrule
\multicolumn{10}{@{}l}{\emph{California smoking}} \\
\quad simultaneous & 2.73 & \textbf{2.66} & 3.14 & 5.27 & 4.07 & 4.37 & 3.85 & 3.29 & 3.13 \\
\quad staggered & \textbf{3.08} & 5.71 & 6.90 & 8.12 & 5.24 & 5.09 & 3.19 & 4.02 & 3.53 \\
\midrule
\multicolumn{10}{@{}l}{\emph{Kansas}} \\
\quad factor model & \textbf{0.014} & 0.019 & 0.031 & 0.079 & 0.051 & 0.064 & 0.016 & 0.041 & 0.210 \\
\bottomrule
\end{tabular}}
\end{table}

\begin{table}[!htbp]
\centering
\small
\caption{MatrixFormer under three panel orientations, in the outcomes' units; lower is better. Standard has units as rows and supplies every main result; transposed has time periods as rows; average takes the mean of the two completed matrices. Entries are means over draws with the designs of a benchmark weighted equally; the lowest value in each column is in bold.}
\label{tab:causal-orientation}
\begin{tabular}{@{}lrrrrrr@{}}
\toprule
 & \multicolumn{3}{c}{Treated-block RMSE} & \multicolumn{3}{c}{Treatment-effect error} \\
\cmidrule(lr){2-4}\cmidrule(l){5-7}
Orientation & California & CPS & Kansas & California & CPS & Kansas \\
\midrule
Standard & 11.70 & 0.532 & \textbf{0.027} & \textbf{2.90} & 0.093 & \textbf{0.014} \\
Transposed & 12.05 & 0.538 & 0.028 & 4.25 & 0.096 & 0.016 \\
Average & \textbf{10.92} & \textbf{0.521} & \textbf{0.027} & 3.11 & \textbf{0.087} & \textbf{0.014} \\
\bottomrule
\end{tabular}
\end{table}

\paragraph{Panel orientation.}
A panel can also be given to MatrixFormer transposed, with time periods as rows: we standardize the columns of the transposed panel, impute, undo the standardization and transpose back. A third variant averages the two completed matrices and requires two forward passes. \Cref{tab:causal-orientation} compares the three on the same draws. The transposed panel is worse on every benchmark and metric. Averaging lowers RMSE on all three benchmarks, and in ten of the twelve designs, but it lowers effect error only on CPS and raises it on California. Every main result uses the standard orientation, rather than an orientation selected from held-out scores.

\FloatBarrier

\FloatBarrier

\section{Recommender-system rating prediction}
\label{sec:recsys-appendix}
\label{sec:recsys-protocol}

We evaluate rating completion on MovieLens 100K \citep{harper2015movielens} and the Netflix small subset distributed by Cornac \citep{salah2020cornac}. These are sparse user--item matrices with ratings on a 1--5 scale. We reuse the benchmark's frozen 80/20 rating-entry holdout, which scores 19,971 ratings on MovieLens 100K, 121,470 on Netflix small. This is an entry holdout, not a held-out-user split.

\paragraph{Metrics.}
We report RMSE and MAE in rating units; lower is better. Following the Cornac harness, each metric is computed within each test user and then averaged equally over users, so the RMSE is the mean of per-user RMSEs rather than a pooled rating-level RMSE.

\paragraph{Methods.}
We compare MatrixFormer with four rating-prediction baselines implemented in Cornac \citep{salah2020cornac}. Matrix factorization with user and item biases \citep{koren2009matrix} and probabilistic matrix factorization (PMF) \citep{salakhutdinov2008pmf} predict a rating from learned low-dimensional user and item vectors; the former is run as Cornac's MF and SVD models, which give identical scores here and are listed once, as Biased MF. UserKNN and ItemKNN are neighborhood methods that predict from the ratings of the most similar users \citep{resnick1994grouplens} or items \citep{sarwar2001item}. The baselines use the same split protocol and are not retuned. MatrixFormer's predictions are clipped to the observed training rating range.
The single model uses observed-only two-way biases and shrunk user scaling, with overlapping windows of at most 2,000 rows and 1,000 columns. Inference settings, scoring definitions, and baseline hyperparameters are given
in \cref{sec:recsys-details}.

\paragraph{Results.}
MatrixFormer obtains RMSE/MAE of 0.8926/0.7330 on MovieLens 100K and 0.8666/0.7364 on Netflix small. It has lower error than every displayed baseline on MovieLens 100K; on Netflix small, Biased MF has lower RMSE/MAE (0.8621/0.7358), while the remaining baselines have higher error.
\Cref{fig:empirical-overview}(b) compares RMSE on the two datasets. The full RMSE and MAE comparison appears in \cref{tab:recsys-ratings} in the appendix. These point-estimate differences are descriptive, not significance claims.

\subsection{Recommender-system evaluation details}
\label{sec:recsys-details}
\begin{figure}[t]
\centering
\includegraphics[width=\linewidth]{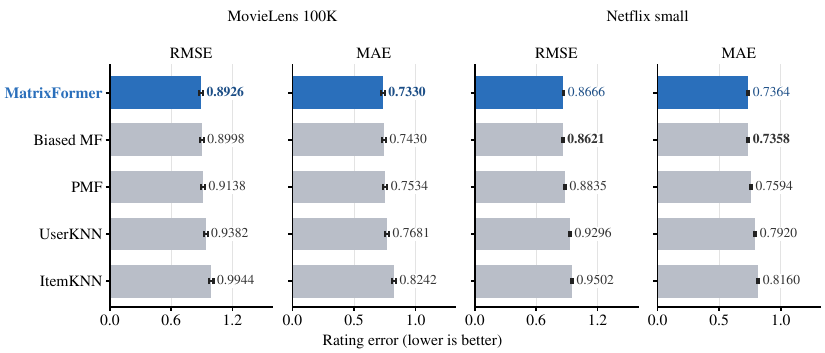}
\caption{Rating prediction in original rating units; lower is better. Bars show RMSE and MAE macro-averaged over test users, with four-decimal labels and the lowest value in each panel in bold. Whiskers on all bars are 95\% test-user bootstrap confidence intervals conditional on the fitted predictors and fixed split. MatrixFormer is blue; baselines are gray. MF and SVD give identical scores and are shown once, as Biased MF.}
\label{tab:recsys-ratings}
\end{figure}

\paragraph{Datasets and splits.}
We use Cornac's MovieLens 100K, Netflix small, MovieLens 1M, and MovieLens
10M datasets \citep{harper2015movielens,salah2020cornac}. The main-text
comparison focuses on MovieLens 100K and Netflix small; the evaluation also
includes the two larger datasets. In the order listed above, the training
matrices have dimensions $943\times1656$,
$9976\times4895$, $6040\times3679$, and $69878\times10643$, with 80,000,
486,242, 800,167, and 8,000,043 observed ratings, respectively.
All training users and items are retained. After unknown-user/item exclusion,
test metrics cover 942, 9,202, 6,038, and 69,797 users, respectively.
The 80/20 rating-entry holdouts use seed 123 and are fixed before the model
and normalization comparisons; users are not held out as a group.

\paragraph{Model inference.}
MatrixFormer uses predictive-mean inference with FP32 weights, BF16 autocast,
and sequential $2000\times1000$ windows with overlapping-tail averaging.
From observed
ratings only, it fits a global mean plus user and item biases with
regularization 3 for 30 alternating iterations. It scales the remaining
residuals globally and by shrunk user scales (prior count 20, clipped to
$[0.5,2]$), and reverses these transformations after inference. The model's
rating predictions are clipped to the observed training range ($[0.5,5]$ for
MovieLens 10M and $[1,5]$ otherwise). No held-out rating enters the
normalization statistics.
For the larger datasets, block inference collects raw predictions only at
observed and held-out coordinates and averages overlapping windows, avoiding
storage of dense completion matrices. Scores are independently recomputed
from predictions.

\paragraph{Scoring.}
For each user $u$ with held-out ratings indexed by $T_u$, let
$e_{ui}=\widehat r_{ui}-r_{ui}$ after inverse normalization and clipping.
For the set $U$ of scored users, we report
\[
\mathrm{MAE}=\frac{1}{|U|}\sum_{u\in U}
  \frac{1}{|T_u|}\sum_{i\in T_u}|e_{ui}|,
\qquad
\mathrm{RMSE}=\frac{1}{|U|}\sum_{u\in U}
  \sqrt{\frac{1}{|T_u|}\sum_{i\in T_u}e_{ui}^{2}}.
\]
No ranking metrics are included. We report neither cross-seed uncertainty
nor a blind-test claim: these public splits have been inspected during research.

\paragraph{Confidence intervals.}
For each saved predictor, we compute each test user's RMSE and MAE before
resampling users with replacement 1,000 times (seed 0). The 2.5th and 97.5th
percentiles of the resulting user-mean scores form the asymmetric 95\%
intervals. The same user resamples are used across methods and metrics on
each dataset. These intervals condition on the fitted models, frozen rating
split, and observed item set; they do not measure training-seed variability
or uncertainty over a new population of items. Saved predictions support
intervals for MatrixFormer and all four displayed baselines on
all four datasets. The MovieLens 100K/Netflix baselines were rerun with their
original settings on the frozen splits to recover per-user errors; neural
predictions were not rerun.

\paragraph{Normalization and context-size controls.}
We compare global and column-wise normalization for an earlier checkpoint of the model
at the same context dimensions as its primary global-normalization runs.
Both use observed training entries only, with statistics computed before
windowing.

A separate context-size control uses an earlier checkpoint, before the matrix sizes were enlarged.
It crosses three user-row limits, 1000, 2000,
and 9976, with global and column-wise normalization on MovieLens 100K and
Netflix small. The item-column limit remained 1000. Those configurations used
identical weights, observations, test cells, and output clipping, with
identical normalization statistics within each normalization setting.
MovieLens 100K predictions agreed within $2\times10^{-6}$ rating units
because all 943 users already fit. On Netflix, increasing the row limit from
2000 to all users improved RMSE/MAE from 0.9341/0.8017 to 0.9070/0.7819
with global normalization and from 0.9330/0.8001 to 0.9058/0.7795 with
column-wise normalization.

\paragraph{Memory and numerical equivalence.}
These measurements also use the earlier checkpoint. Streamed inference
microbatches the encoder, independent attention sequences, feed-forward cells,
and output-distribution decoding. It retains the complete attended row or
column sequence, FP32 weights, BF16 autocast, and predictive-mean calculation.
On a $2000\times1000$ training-only Netflix block, maximum prediction deviation
from dense execution is $1.05\times10^{-6}$ rating units, with RMS deviation
$4.62\times10^{-8}$. Model weights remain fixed, and no gradients are computed.
Peak allocated GPU memory falls from 93.5 to 6.9 GiB.
Training-only capacity probes at 4000, 8000, and all 9976 users require
13.3, 26.3, and 32.7 GiB, respectively, at 1000 items on the same H200.
The largest user context in this control was selected by memory
feasibility, not test error. The full Netflix rating evaluation takes
approximately 133 seconds per normalization versus approximately 89 seconds
for the earlier dense 2000-row configuration; this is not an equal-runtime
comparison.

\paragraph{Baseline settings and score verification.}
The five Cornac rating-prediction implementations are MF, PMF, SVD,
UserKNN, and ItemKNN; identical MF/SVD scores are displayed once, as Biased MF.
MF, PMF, and SVD use rank 10, with 25, 100,
and 25 iterations, respectively. Learning rates are 0.01, 0.001, and 0.01;
regularization parameters are 0.02, 0.001, and 0.02.
MF uses biases; both nearest-neighbor baselines use 20 neighbors, Pearson
similarity, and mean centering. All use seed 123.
SVD, PMF, UserKNN, and ItemKNN were rerun on MovieLens 100K and Netflix small
using Cornac 2.3.5, preserving the frozen train/test indices and training-row
order. Per-entry predictions are saved and independently rescored against
Cornac's per-user metrics. All five implementations are also evaluated on the fixed MovieLens
1M/10M splits, where their predictions are independently rescored at full precision.

\begin{figure}[t]
\centering
\includegraphics[width=\linewidth]{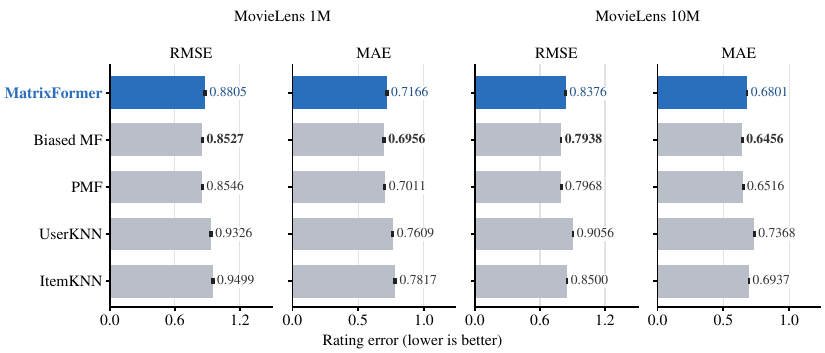}
\caption{Rating prediction on MovieLens 1M and 10M in original rating units; lower is better. RMSE and MAE are macro-averaged over test users; whiskers are asymmetric 95\% test-user bootstrap confidence intervals conditional on fitted predictors and the fixed split. Blue bars show MatrixFormer; gray bars show the same baselines as the main-text comparison. MF and SVD give identical scores and are shown once, as Biased MF. Bold labels mark the lowest displayed error in each panel.}
\label{fig:recsys-large}
\end{figure}

\FloatBarrier

\section{Benchmark-score completion for language-model evaluation}
\label{sec:llm-appendix}
\label{sec:llm-protocol}
\label{sec:llm-results}

Evaluating a language model on every benchmark is expensive. We study whether MatrixFormer can reconstruct a model's unmeasured benchmark scores after observing a small subset of its evaluations. Rows represent models and columns represent benchmark tasks; this is score-matrix completion, not a new evaluation of generated answers. We use the released matrices and benchmark-selection algorithms of \citet{smola2026submodular}, summarized in \cref{tab:llm-datasets}: MTEB, English-task leaderboard scores for embedding models \citep{muennighoff2023mteb}; MMLU, per-subject leaderboard scores on its 57 subjects \citep{hendrycks2021mmlu}; Merged, a heterogeneous matrix assembled from nine collections; and BenchPress, the curated model--benchmark matrix of \citet{benchpress2026matrix}, which here names a dataset rather than an imputation baseline. All are fixed snapshots, not live leaderboards.

\begin{table}[!htbp]
\centering
\small
\caption{Released benchmark-score matrices used for evaluation. Observed
entries are the available scores before holding out model rows.}
\label{tab:llm-datasets}
\begin{tabular}{@{}lrrr@{}}
\toprule
Dataset & Models & Tasks & Observed \\
\midrule
MTEB & 263 & 56 & 77.3\% \\
MMLU & 5,452 & 57 & 100.0\% \\
Merged & 118 & 114 & 31.1\% \\
BenchPress & 83 & 49 & 33.8\% \\
\bottomrule
\end{tabular}
\end{table}

\paragraph{Protocol.}
Each matrix is split into ten folds of model rows, holding out approximately 10\% of the models per fold. The remaining models form the context, capped at 3,800 by a deterministic subsample; the cap affects only MMLU. For each held-out model we reveal its available scores on $k$ selected tasks, for $k=1,\ldots,15$, and predict its other observed scores. Naturally missing entries are not ground truth, so on the sparse matrices $k$ selected tasks may reveal fewer than $k$ scores.

\paragraph{Task selection.}
Which tasks are revealed is decided by one of three rules from the release of \citet{smola2026submodular}. Each chooses tasks one at a time, and the two greedy rules use the task-correlation matrix estimated from the context models. Random selection draws the tasks uniformly at random. Greedy entropy adds the task with the largest conditional variance given the tasks already selected, that is, the task hardest to predict from them. Greedy mutual information (MI) adds the task with the largest mutual-information gain between the selected and the remaining tasks, that is, the task most informative about those not yet selected. The selection rule is a setting of the experiment, not a method under comparison: task orders are computed once per fold, and every predictor receives the same revealed scores.

\paragraph{Methods.}
MatrixFormer receives the context models and the revealed scores of the held-out models as one partially observed matrix, and its predictive mean fills the unrevealed entries; no weights are updated. We compare it with the predictor of the selection release, which we label kernel ridge. Let $R$ be the task-correlation matrix estimated from the context, $O$ the tasks revealed for a held-out model and $U$ its other tasks. In standardized coordinates the prediction is
\[
\hat z_U = R_{UO}(R_{OO}+0.01 I)^{-1}z_O,
\]
the ridge-regularized conditional mean of a Gaussian with correlation $R$. This is kernel ridge regression on task indices with kernel matrix $R$, not an RBF kernel. As in the release, the revealed inputs $z_O$ are clipped to $[-10,10]$, and a model with no revealed score receives the context mean.

\paragraph{Metric.}
We report the root mean squared error in standardized units. Let $\mathcal{E}$ be the entries of a fold's held-out models that are observed and not revealed, and let $z_{ij}$ and $\hat z_{ij}$ be the target and the prediction, standardized with the context mean and standard deviation of task $j$ and clipped to $[-10,10]$. Then
\[
\mathrm{RMSE}=\Bigl(\frac{1}{|\mathcal{E}|}\sum_{(i,j)\in\mathcal{E}}(z_{ij}-\hat z_{ij})^2\Bigr)^{1/2},
\]
averaged over the ten folds; lower is better. One unit is a task's standard deviation over the context models. Every held-out model is scored, including those with no revealed score, and all predictors share the same context, revealed scores and target entries.

\begin{figure}[!htbp]
\centering
\includegraphics[width=\linewidth]{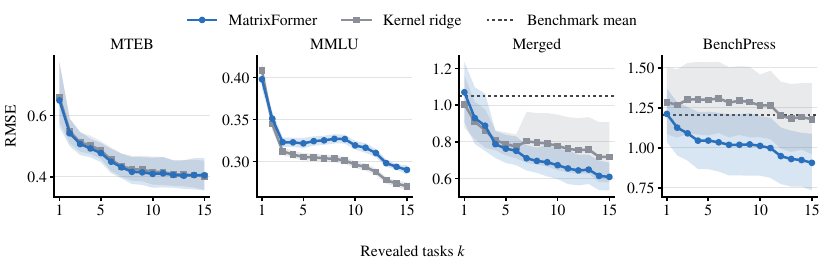}
\caption{\textbf{Benchmark-score completion under greedy MI selection.} Standardized RMSE on the unrevealed scores of held-out models against the number of revealed tasks $k$; lower is better. Lines are means over ten folds and bands are $95\%$ bootstrap intervals over the folds. The dashed line is the error of predicting each task's context mean; on MTEB and MMLU it lies above the plotted range.}
\label{fig:llm-rmse-mi}
\end{figure}

\begin{figure}[!htbp]
\centering
\includegraphics[width=\linewidth]{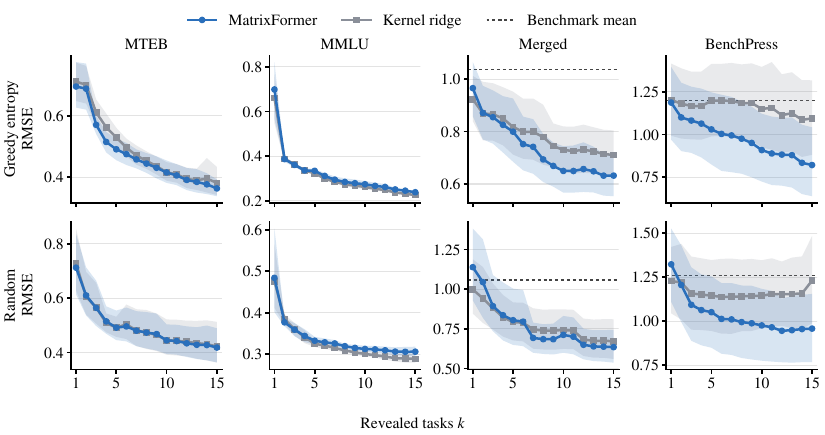}
\caption{\textbf{Benchmark-score completion under greedy entropy and random selection.} Same quantities as \cref{fig:llm-rmse-mi}. Random selection uses one order per fold.}
\label{fig:llm-rmse-entropy-random}
\end{figure}

\paragraph{Results.}
\Cref{fig:llm-rmse-mi} shows the error under greedy MI selection. On the two sparse matrices MatrixFormer has the lower error: at $k=15$ its RMSE is $0.608$ against $0.717$ on Merged and $0.907$ against $1.179$ on BenchPress, where kernel ridge stays close to the error of predicting each task's context mean. At $k=5$ the margin on Merged is small ($0.763$ against $0.788$). On the two dense matrices the methods are close: on MTEB they are within each other's intervals at every budget ($0.405$ against $0.401$ at $k=15$), and on MMLU kernel ridge is lower for $k\ge2$ ($0.271$ against $0.290$ at $k=15$). Greedy entropy and random selection give the same pattern (\cref{fig:llm-rmse-entropy-random}). Across all 180 combinations of dataset, selection rule and budget, MatrixFormer has the lower fold-mean RMSE in 123: 44 of 45 on BenchPress, 40 on MTEB, 34 on Merged and 5 on MMLU. These counts summarize correlated settings, not independent tests.

On Merged under greedy MI, the kernel-ridge interval widens at $k=7$ because of a single fold. The seventh task selected in that fold has only two context scores, and one held-out model's score on it is $15$ context standard deviations above the mean; kernel ridge's RMSE on the fold rises to $1.41$, against $0.72$ for MatrixFormer. Without this fold MatrixFormer remains lower ($0.602$ against $0.637$ at $k=15$), by a smaller margin. The release also provides a logit-space variant of kernel ridge. Given the same greedy-MI tasks on Merged and BenchPress, its RMSE at $k=15$ is $0.714$ and $1.312$, above MatrixFormer on both.

\paragraph{Intervals and relation to the source paper.}
The intervals resample the ten fold-level scores 10,000 times. Contexts overlap between folds, so they describe variation across folds under the fixed protocol and are not significance tests. Our protocol differs from the released evaluation in its scoring, not in its selection algorithms: the released scorer skips held-out models with no revealed score and does not clip predictions, whereas we score every held-out model and apply the same clipping to all predictors. Fixing the task orders before inference also ensures that every predictor is given the same tasks when selection is sensitive to numerical ties, as on Merged.

\FloatBarrier

\FloatBarrier

\section{Representation experiments: construction and scope}
\label{sec:probe-setup}

\subsection{Latent-factor experiment on the CLS tokens}
\label{sec:cls-probe-details}
\begin{figure}[!htbp]
\centering
\includegraphics[width=\linewidth]{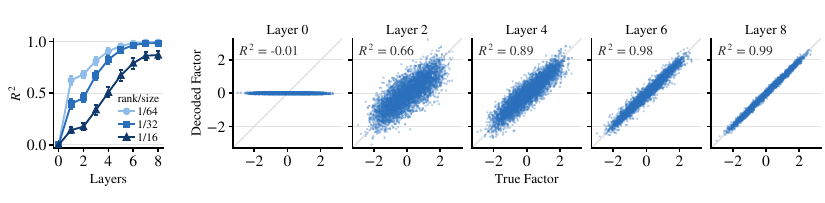}
\caption{\textbf{Latent column factors in the column-CLS tokens across layers.}
Same matrices and linear projection as \cref{fig:cls-factors}(b), for the column-CLS tokens.
Left: out-of-fold $R^2$ of the linear projection by layer; each
point averages 30 matrices with $95\%$ intervals. Right: true against decoded
column factors at rank/size $1/64$, pooling all factors of one matrix each of
size 128, 256 and 512 (ranks 2, 4 and 8). The true column factor is $V_j/s_j$,
the column factor after column standardization; the panel value is $R^2$
averaged over these factors.}
\label{fig:cls-factors-columns}
\end{figure}

We generate noiseless matrices $X=UV^\top$ with standard-Gaussian
$U,V\in\mathbb{R}^{n\times r}$ and observe $40\%$ of entries uniformly at random,
for rank-to-size ratios $r/n\in\{1/64,1/32,1/16\}$ and three sizes per ratio,
ten matrices each. After layer $\ell$, we concatenate the $K_c$ row-CLS tokens of
row $i$ into $c_i^{(\ell)}\in\mathbb{R}^{K_cd}$ and regress $U_i$ on
$c_i^{(\ell)}$ by ridge regression, fitting on seven eighths of the rows and
scoring $R^2$ on the held-out eighth, over eight folds. At layer $0$ the tokens
are identical across rows. Recovery declines with the absolute rank:
at $r/n=1/16$, $R^2$ falls from $0.93$ at $n=64$ to $0.73$ at $n=256$.
The same experiment on the $K_r$ column-CLS tokens of each column recovers the
column factors to the same degree ($R^2=0.99$, $0.98$ and $0.87$;
\cref{fig:cls-factors-columns}).
MatrixFormer standardizes each column of its input, but not each row, by the
standard deviation $s_j$ of that column's observed entries. The model therefore
sees $x_{ij}/s_j=U_i^\top(V_j/s_j)$: the row factors are unchanged, and the
column target is $V_j/s_j$. Since $s_j$ depends only on column $j$'s own
observed entries, it carries no information from the other columns.

As a check on the eight-fold score, we set aside one ninth of the rows, fit the
feature scaling, the ridge penalty and the regression on the remaining rows
only, and score the untouched ninth. At layer 8 this gives $R^2=0.99$, $0.98$
and $0.87$ for the row factors and $0.99$, $0.98$ and $0.87$ for the column
factors, matching the eight-fold values. Permuting the factors across rows
before fitting gives $R^2$ between $-0.03$ and $-0.01$, so the scores do not
come from the dimension of the token vector exceeding the number of rows.

In \cref{fig:cls-factors}(a), the example is a rank-2 matrix of size 128,
masked to $40\%$ observed entries; $\widehat U$ is the linear projection of each
row's CLS tokens at layer 8, shown beside the true $U$.
For display only, the columns of the matrix are ordered by their first factor;
the rows keep their original order.
In \cref{fig:cls-factors}(b), each point of the left panel averages 30 matrices
(three sizes, ten each) with $95\%$ intervals.
Layer $\ell$ is the output of transformer block $\ell$; layer $0$ is the encoder
output. The right panels pool all factors of one matrix each of size 128, 256
and 512 (ranks 2, 4 and 8); the panel value is $R^2$ averaged over these factors.

\subsection{Latent-group experiment on the cell embeddings}
\label{sec:group-probe-details}

\paragraph{Example in the main text.}
The matrix in \cref{fig:synthetic-embedding-example} has four row groups of 16
rows each. Each group has a profile of 64 independent standard-normal entries;
every row equals its group's profile plus independent Gaussian noise of
standard deviation $0.25$. Rows are shuffled and $40\%$ of the entries are
observed uniformly at random. The seed and condition were fixed before
inference. Let $h_{ij}^{(\ell)}\in\mathbb{R}^{768}$ be the embedding of cell
$(i,j)$ after layer $\ell$, with the CLS border removed; layer $0$ is the
encoder output. The row embedding averages the row's cells, observed and masked:
\[
z_i^{(\ell)}=\frac{1}{C}\sum_{j=1}^{C}h_{ij}^{(\ell)}.
\]
$K$-means with $K=4$ and 20 initializations clusters the full 768-dimensional
row embeddings; the number of groups is supplied, the memberships are not.
The adjusted Rand index (ARI) compares the clusters with the true groups: it is
one for perfect recovery, approximately zero for chance agreement, and can be
negative. PCA is used only for display.

\FloatBarrier

\end{document}